\documentclass[journal]{IEEEtran}

\usepackage{cite}
\usepackage[utf8]{inputenc}
\usepackage{graphicx}
\usepackage{amsmath}
\usepackage{amssymb}
\usepackage{amsthm}
\usepackage{booktabs}
\usepackage{multirow}
\usepackage[table]{xcolor}
\usepackage{array}
\usepackage{makecell}
\usepackage{siunitx}
\usepackage{enumitem}
\usepackage{soul}
\usepackage{url}
\usepackage{float}
\usepackage{caption}
\usepackage[hidelinks]{hyperref}

\setlist[itemize]{itemsep=0pt, parsep=0pt, topsep=2pt}
\newcolumntype{M}[1]{>{\centering\arraybackslash}m{#1}}

\title{AERIS: Aerial-Edge Role-Driven Intelligence at Runtime via Orchestrated Language-Model Swarms}

\author{Jiabin~Lou, Haopeng~Wang, Xinyu~Liu, Yu~Zhang, Rongye~Shi, and Wenjun~Wu%
\thanks{This work has been submitted to the IEEE for possible publication. Copyright may be transferred without notice, after which this version may no longer be accessible. This work was supported in part by the National Key Research and Development Program of China under Grant 2025YFF1505704, in part by the State Key Laboratory of Complex and Critical Software Environment under Grant SKLCCSE, in part by the National Natural Science Foundation of China under Grant 62306023, and in part by the Beijing Nova Program under Grant 20240484490. \textit{Corresponding author: Wenjun Wu.}}
\thanks{Jiabin Lou is with the School of Computer Science and Engineering, Beihang University, Beijing 100191, China, and also with Hangzhou International Innovation Institute, Beihang University, Hangzhou 311115, China (e-mail: loujiabin@buaa.edu.cn).}%
\thanks{Haopeng Wang, Rongye Shi, and Wenjun Wu are with the School of Artificial Intelligence, Beihang University, Beijing 100191, China, and also with Hangzhou International Innovation Institute, Beihang University, Hangzhou 311115, China (e-mails: \{by2442216, shirongye, wwj09315\}@buaa.edu.cn).}%
\thanks{Xinyu Liu and Yu Zhang are with the School of Artificial Intelligence, Beihang University, Beijing 100191, China (e-mails: \{liuxy0109, stevezhang\}@buaa.edu.cn).}%
}

\begin{document}

\maketitle

\begin{abstract}
Integrating large language models into robotic systems holds promise for enhancing autonomy, yet practical deployment remains constrained by strict heartbeat-constrained scheduling and limited computational power. We propose AERIS: an edge deployment framework for aerial platforms. It organizes dedicated small language models combined with lightweight perception and control modules into roles that can be instantiated at runtime, and dynamically rebinds them across different executors as resources change, thereby pushing intelligent capabilities to the edge. AERIS achieves long-horizon instruction decomposition through an attention–subgoal alignment mechanism, which involves annotating the currently active instruction step in messages, thereby progressively approaching long-term objectives. We evaluate AERIS on a high-fidelity UAV Vision-and-Language Navigation benchmark. Under a heartbeat-timed execution mechanism, AERIS maintains a stable perception–decision–control loop between a low-frequency planner and a high-frequency controller, supporting real-time closed-loop operation. We further validate its deployability through two real-world experiments focused on planning and fast response. A demonstration video is provided in the supplementary materials.
\end{abstract}


 \begin{figure*}[htbp]
\centering
\includegraphics[width=0.8\linewidth]{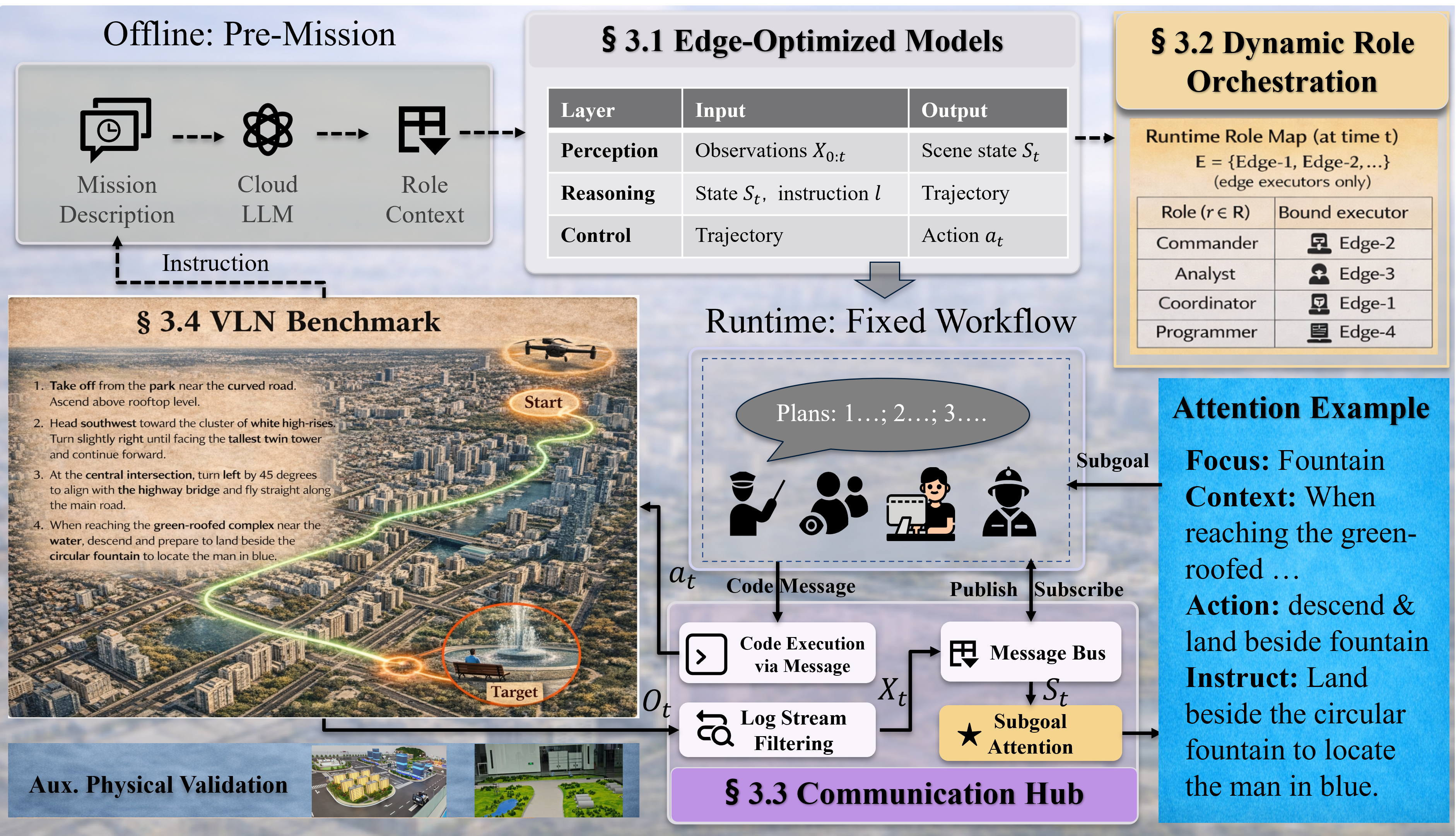}
\caption{\textbf{Overview of AERIS.}
The perception layer converts onboard observations into a typed state $S_t$; the semantic layer outputs a schema-constrained decision $D_t$. A Communication Hub routes and validates messages, performs instruction-step attention--subgoal alignment (ATT), and binds $D_t$ into executable control commands. An orchestration engine instantiates roles on heterogeneous edge executors and updates role--executor bindings at runtime.
}
\label{fig:edge}
\end{figure*}

\section{Introduction}
\label{Introduction}
Large language models (LLMs) have shown promising potential in robotics, such as understanding natural-language instructions, reasoning about current states, and adapting to previously unseen environments~\cite{monwilliams2025embodied}. Compared with conventional robotic systems that rely on fixed task interfaces or predefined rules, language models provide a more natural form of human--robot interaction and enable robots to make use of richer semantic information for complex decision making. However, deploying these capabilities in real robotic systems remains challenging. First, robots operate within a strict perception--decision--control loop, and additional latency introduced by high-level reasoning may disrupt synchronization with low-level control, thereby affecting operational safety. Second, LLM inference is computationally expensive, making long-term stable deployment difficult on resource-constrained platforms. Third, robots often operate under unstable wireless connectivity, and continuous reliance on cloud models can further weaken real-time responsiveness and reliability.\par
These challenges are particularly pronounced in aerial missions~\cite{10776775}. For example, in inspection or search-and-rescue scenarios, an operator may issue a multi-step natural-language instruction such as ``fly along the road, turn left after passing the red-roofed building, and hover near the bridge.'' To complete such tasks, a UAV must not only interpret landmarks, action order, and intermediate objectives in the instruction, but also determine the currently relevant instruction content from continuously changing onboard visual observations and promptly convert high-level semantic outputs into executable control commands. Such instructions usually consist of a sequence of local actions and landmark constraints, while the UAV may encounter viewpoint changes, occlusions, and environmental variations during flight. It therefore needs to infer task progress from ongoing observations and adjust subsequent motion accordingly. Consequently, long-horizon UAV navigation depends not only on natural-language understanding, but also on stable coupling between semantic decision making and low-level control in dynamic environments.\par
Existing approaches often address only part of these requirements. Cloud-hosted or large-scale models can provide strong semantic understanding and reasoning capabilities, but they are difficult to use reliably for real-time closed-loop operation under communication fluctuation, limited computation, and strict timing constraints. Lightweight end-to-end vision--language policies can be efficient at inference time, yet their intermediate decisions are usually implicit in latent representations, making it difficult to expose clear interfaces for runtime validation, modular coordination, and adaptive resource scheduling~\cite{das2026latencyaware}. For aerial-edge systems, a key challenge is therefore to organize natural-language instruction understanding, visual environment perception, high-level semantic decision making, and low-level flight control into a continuously operating system under constrained resources and real-time requirements. The semantic layer must maintain valid decisions as the environment evolves, the control layer must preserve flight stability at a higher frequency, and the overall system must remain responsive under dynamically changing edge resources.\par
To address these challenges, this paper introduces \textbf{A}erial-\textbf{E}dge \textbf{R}ole-driven \textbf{I}ntelligence at runtime via orchestrated language-model \textbf{S}warms ({AERIS}), a role-driven framework for aerial-edge autonomy. AERIS organizes specialized small language models (SLMs), lightweight perception modules, and control modules into runtime roles, coordinates their interactions through schema-constrained message passing, and employs an attention--subgoal alignment mechanism to track the currently active instruction step. To cope with dynamic edge resources, AERIS maintains a runtime role mapping table and dynamically rebinds roles across executors according to load and latency while preserving the high-level workflow. In this way, AERIS integrates natural-language understanding, visual landmark perception, semantic decision making, and high-rate control within a unified runtime framework, enabling language-driven autonomy to operate more reliably on resource-constrained UAV platforms. We evaluate AERIS on long-horizon UAV Vision-and-Language Navigation tasks to examine its joint capability in semantic understanding, landmark grounding, and real-time closed-loop execution, and further conduct real-world UAV experiments to demonstrate deployability under practical constraints.\par

\par
In summary, our contributions are:
\begin{itemize}
    \item We propose {AERIS}, a role-driven aerial-edge framework that decomposes LLM-based autonomy into specialized runtime roles with hub-mediated, schema-constrained communication and stepwise grounding;
    \item We develop a dynamic orchestration mechanism with runtime re-binding to maintain responsiveness under compute and network variability; and
    \item We present a comprehensive evaluation protocol that jointly reports navigation performance and system-level reliability, complemented by real-world validation.
\end{itemize}

\section{Related Works}
\subsection{LLM-Enabled Embodied Robotics}
LLMs have been increasingly used as high-level decision makers that translate natural-language goals into executable action sequences for robots.
Early systems couple LLM planning with affordance checking or tool execution, enabling grounded task completion under open-vocabulary instructions (e.g., SayCan~\cite{ahn2022do}, Code-as-Policies~\cite{liang2022codeaspolicies}).
Beyond purely symbolic planning, vision-language-action (VLA) and embodied foundation models integrate multimodal perception and action prediction to improve generalization and robustness, such as RT-2~\cite{zitkovich23a}.
More recent works further push towards generalist, reusable VLA policies and open ecosystems, including Octo~\cite{ghosh2024octo}, supported by large-scale real-robot datasets such as DROID~\cite{khazatsky2024droid}.
In parallel, efficiency-aware VLA designs have emerged to better match on-robot resource budgets, e.g., dynamic inference mechanisms in DeeR-VLA~\cite{yue2024deervla}.
Despite this progress, most prior work still targets planning with relatively loose control rates or relies on substantial compute; addressing hard real-time constraints and stable closed-loop execution on resource-limited platforms remains less explored.

\subsection{Language model-based Aerial Autonomy}
Language model-based aerial autonomous systems also introduce additional challenges, such as long-range navigation planning in large-scale outdoor environments, high-complexity three-dimensional spatial decision-making, and perception uncertainties caused by rapid motion. In this area, AerialVLN~\cite{liu2023aerialvln} establishes a city-scale VLN benchmark, extending continuous VLN evaluation from ground agents to aerial. Subsequent efforts, such as CityNavAgent~\cite{zhang2025citynavagent} and OPENUAV~\cite{wang2024openuav}, further explore aerial navigation with stronger world modeling or agentic reasoning.
At the visual grounding level, related multimedia research has also explored region-level language grounding, such as CLIP-based adaptation for visual grounding \cite{xiao2024clipvg}, which is complementary to instruction-following navigation but does not address closed-loop aerial execution.
However, existing work largely emphasizes planning ability and is mostly evaluated in simulation, while methods designed for edge execution under tight latency budgets remain relatively unexplored.

\subsection{Edge-Optimized Language Models}
Deploying LLM capabilities on the edge typically requires systematic compression and inference acceleration to meet tight memory and latency budgets.
Recent post-training quantization methods enable low-bit inference while preserving generation quality (e.g., SpQR~\cite{dettmers2024spqr}, SpinQuant~\cite{liu2024spinquant}).
For long-context workloads, the KV cache often becomes the dominant memory bottleneck, and techniques such as KIVI~\cite{liu2024kivi} reduce cache footprint with tuning-free low-bit quantization.
At the kernel level, attention implementations improve hardware utilization and reduce memory traffic, e.g., FlashAttention-3~\cite{shah2024flashattention3}.
At the decoding level, multi-token generation frameworks such as Medusa~\cite{cai2024medusa} reduce per-token latency.
For embodied autonomy, these techniques are most effective when co-designed with a system architecture that explicitly budgets computation across perception, reasoning, and control.
Our approach follows this principle by designing an edge-oriented multi-scale model stack and orchestration mechanism that can adaptively trade accuracy for timeliness during runtime.

\subsection{LLM-Based Multi-Agent Systems}
LLM-based multi-agent systems decompose complex problems into multiple interactive agents with specialized roles, enabling collaborative reasoning, planning, reflection, and task execution through language-mediated communication.
Representative frameworks such as CAMEL~\cite{li2023camel} and AutoAgents~\cite{zhang2024autoagents} show that role specialization and agent-to-agent dialogue can support complex problem solving more effectively than a single monolithic language model.
In particular, AutoAgents moves from predefined agents toward task-adaptive agent generation, where expert agents are dynamically instantiated and coordinated according to the current task.
Recent studies further shift the focus from individual agent capabilities to role construction, communication structures, and adaptive collaboration mechanisms in LLM-based multi-agent systems: ProAgent~\cite{zhang2024proagent} uses LLMs to infer teammates' intentions and adapt cooperative behavior, MacNet~\cite{qian2025macnet} studies large-scale agent collaboration through structured interaction topologies, and COLLAB-LLM~\cite{albaroudi2026collabllm} strengthens scalable multi-agent collaboration from the perspectives of communication protocols, hierarchical role architectures, and dynamic task graphs.
These studies suggest that the key advantage of LLM-based multi-agent systems lies not only in task decomposition, but also in exposing interpretable intermediate roles and communication traces that can support inspection, debugging, and iterative refinement.

LLM-based multi-agent coordination has also been extended to embodied intelligence and multi-robot systems.
RoCo~\cite{mandi2024roco} enables multiple robots to discuss task strategies, generate subtask plans, and refine waypoint-level actions using environmental feedback.
Other studies compare centralized, decentralized, and hybrid LLM planning structures for scalable multi-robot collaboration~\cite{chen2024scalable}, and language-model-based agents have also been used to automate multi-agent reinforcement learning pipelines for drone swarm policy training, as in Agents Trainer~\cite{lou2026agentstrainer}.
However, most existing LLM-based multi-agent and multi-robot systems still focus on high-level task reasoning, simulation-level coordination, or offline policy generation. Their ``roles'' are usually logical participants in the reasoning process rather than runtime units under strict timing constraints. Therefore, real-world deployment of LLMs in multi-robot systems remains limited by latency, communication robustness, scalability, and environmental uncertainty.

\section{Methodology}

AERIS is an aerial-edge role-driven intelligence platform that orchestrates language-model swarms at runtime.
As shown in Fig.~\ref{fig:edge}, AERIS integrates three key components:
\begin{itemize}[leftmargin=1.2em, itemsep=2pt, topsep=2pt, parsep=0pt]
  \item \textbf{Edge-optimized model stack} for low-latency UAV perception, decision and control (see Sec.~\ref{sec:model_stack}),
  \item \textbf{Dynamic Role Orchestration} that maintains a runtime role-executor map adaptively under resource variation (see Sec.~\ref{sec:role_orch}), and
  \item \textbf{Communication Hub} which performs instruction-aware message arbitration and subgoal alignment (see Sec.~\ref{sec:comm_hub}).
\end{itemize}
Together, these components enable the stable closed-loop execution of AERIS at the aerial edge and we conduct experiments primarily on a closed-loop UAV VLN benchmark within a runtime decision cycle (see Sec.~\ref{sec:benchmarks}).\par

\begin{figure}[h!]
\centering
\includegraphics[width=0.7\linewidth]{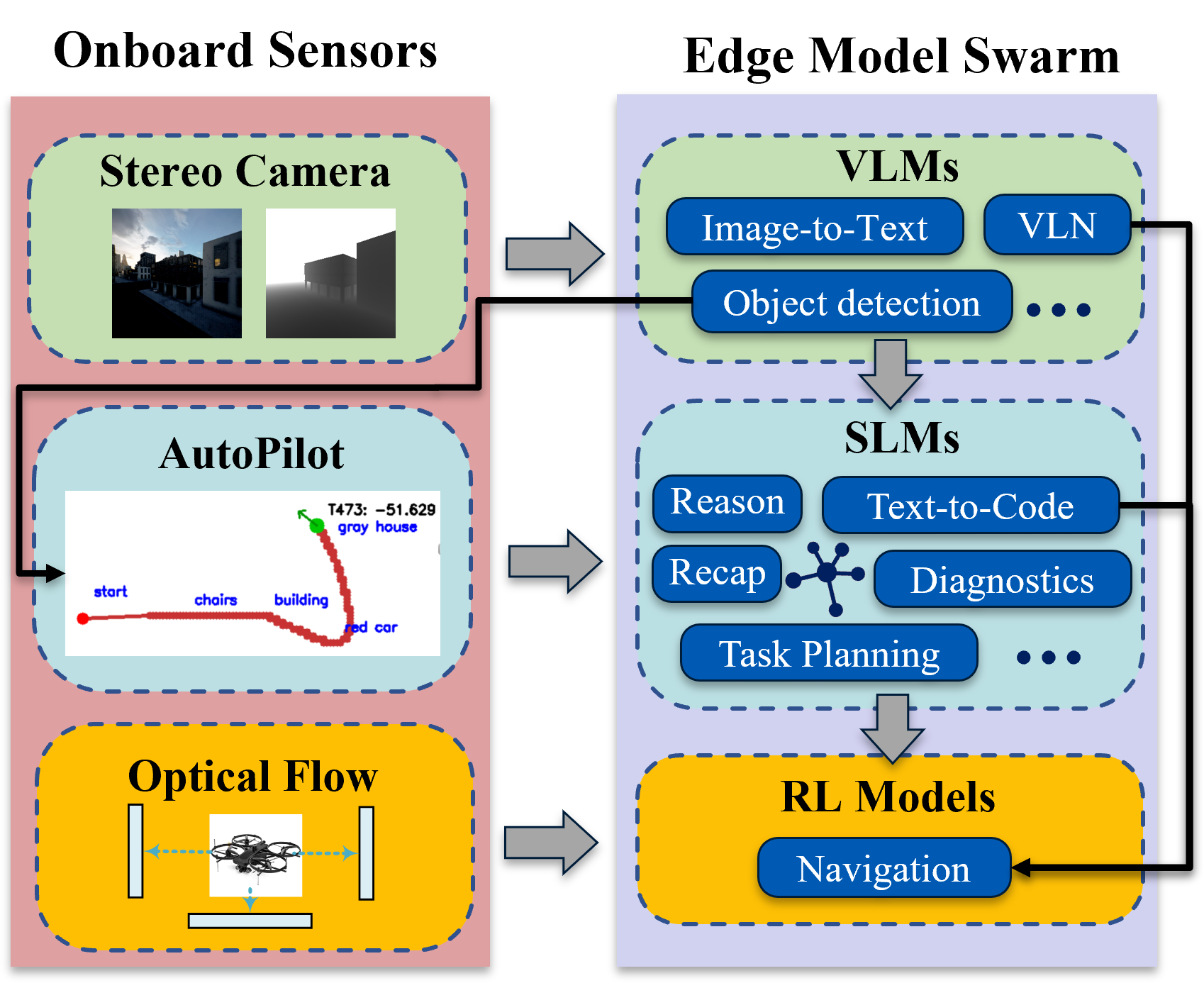}
\caption{\textbf{Edge-optimized model stack in AERIS.}
Perception maps observation history $X_{0:t}$ to a typed state $S_t=\langle\bar S_t, I_t, C_t, O_t\rangle$; semantic reasoning $R_\psi$ maps $(S_t, G, M_{t-1})$ to a schema-constrained decision $D_t$; a high-rate controller executes $D_t$ between semantic heartbeats. }
\label{fig:models}
\end{figure}

\subsection{Edge-Optimized Model Stack}
\label{sec:model_stack}
To satisfy the latency requirements of perception and control in aerial-edge scenarios, AERIS adopts a three-layer model stack (Fig.~\ref{fig:models}). Each layer is specialized and can be instantiated with compact models to trade off latency and accuracy.

\vspace{0.1cm}
\noindent \textbf{Perception Layer.} After the UAV acquires observational data $X_{0:t}$(e.g., FPV RGB image sequences), the perception layer encodes these data into a structured state with comprehensive information, defined by the following equation:
\begin{equation} \phi_{\theta}:\; X_{0:t}\ \mapsto\ S_t, \qquad S_t := \big(\bar{S}_t, I_t, C_t, O_t\big), \end{equation}
Here, $\bar{S}_t$ denotes the global semantic feature summary, $I_t$ captures instance-level semantic information, $C_t$ represents task-relevant key cues, and $O_t$ corresponds to the volumetric voxel occupancy state. To keep this perception--decision interface explicit, Table~\ref{tab:state_contract} summarizes the minimum typed contract consumed by the semantic layer. The contract fixes the required fields, data types, and bounded representations, allowing different perception variants to be exchanged without changing downstream parsing or control binding.

\begin{table}[t]
\centering
\small
\setlength{\tabcolsep}{5pt}
\renewcommand{\arraystretch}{1.06}
\caption{Minimal typed state contract $S_t$ consumed by the semantic layer.}
\label{tab:state_contract}
\begin{tabular}{@{}llll@{}}
\toprule
\textbf{Block} & \textbf{Type} & \textbf{Contents} & \textbf{Bound} \\
\midrule
$\bar{S}_t$ & Text & Scene summary & Max length \\
$I_t$ & List & Top-$k$ detections & $k$ \\
$C_t$ & List & Top-$k$ instruction cues & $k$ \\
$O_t$ & Grid & Ego-frame occupancy & Fixed grid \\
\bottomrule
\end{tabular}
\end{table}

As illustrated in Table~\ref{tab:perception_pool_compact}, we establish a pool of lightweight models containing three variants for each perception function. This design enables real-time deployment under heterogeneous edge computing resource environments. Detailed perception-model acquisition, training, and compression schemes are provided in Appendix~A.

\begin{table*}[h]
\centering
\caption{Perception layer model pool used to construct the typed state $S_t$.}
\label{tab:perception_pool_compact}
\setlength{\tabcolsep}{3pt}
\renewcommand{\arraystretch}{1.12}
\small
\begin{tabular}{M{2.8cm} M{3.4cm} M{3.4cm} M{3.4cm} M{3.0cm}}
\toprule
\textbf{Function} & \textbf{Model A} & \textbf{Model B} & \textbf{Model C} & \textbf{Output} \\
\midrule
Image-to-Text &
OpenELM\mbox{-}0.89B &
Gemma\mbox{-}2.4B &
Phi\mbox{-}2\mbox{-}3.1B &
Caption \\
Object detection &
YOLOv8\mbox{-}n\cite{yolov8_docs} &
RT\mbox{-}DETR R50\cite{rtdetr_cvpr24} &
PP\mbox{-}YOLOE\mbox{-}s\cite{ppyoloe_arxiv} &
Class \& Bbox \\
VLN cue extraction$^{\dagger}$ &
Qwen2.5\mbox{-}0.5B &
Gemma\mbox{-}2.4B &
Phi\mbox{-}2\mbox{-}3.1B &
keypoints \\
3D occupancy &
OccWorld\cite{occworld_eccv24} &
NeMo\cite{nemo_eccv24} &
OCC\mbox{-}VO\cite{occ_vo_2024} &
Occupancy Grid\\
\bottomrule
\end{tabular}

\vspace{2pt}
% \begin{flushleft}
\footnotesize\textit{Notes.}
Common input for all rows is RGB frames/sequences. 
$\dagger$~VLN additionally requires natural-language instructions. \par
% $\ddagger$~3D occupancy/world models typically use surround-view multi-camera and temporal context. \par
The Image-to-Text and VLN rows use models trained with the TinyLLaVA Factory~\cite{jia2024tinyllavafactory} framework; the family prefix is omitted in cells for brevity.
% \end{flushleft}
\end{table*}

\vspace{0.1cm}
\noindent \textbf{Semantic Reasoning Layer.} The reasoning layer consumes the typed state $S_t$, mission context $G$, and a short-horizon semantic memory $\mathcal{M}_{t-1}$, and outputs a structured decision:
\begin{equation} \label{eq:reason_map} \mathcal{R}_{\psi}:\ (S_t,\,G,\,\mathcal{M}_{t-1}) \ \mapsto\ D_t . \end{equation}
$D_t$ is decoded in a typed, schema-constrained format, which makes the interface to downstream execution stable and verifiable. Table~\ref{tab:decision_schema} lists the minimal decision fields consumed by the Communication Hub and controller. We instantiate $\mathcal{R}_{\psi}$ with compact instruction-tuned SLM variants consistent with the model pool in Table~\ref{tab:perception_pool_compact}; all variants share the same schema and are trained with supervised instruction tuning, parameter-efficient adaptation~\cite{hu2022lora,dettmers2023qlora}, and post-training quantization for edge deployment. Detailed semantic training and alignment objectives are provided in Appendix~B.

\begin{table}[t]
\centering
\small
\setlength{\tabcolsep}{4pt}
\renewcommand{\arraystretch}{1.06}
\caption{Minimal decision schema $D_t$ consumed by the hub and controller.}
\label{tab:decision_schema}
\begin{tabular}{@{}llll@{}}
\toprule
\textbf{Field} & \textbf{Type} & \textbf{Meaning} & \textbf{Constraint} \\
\midrule
$a_t$ & Enum & Action mode & Fixed set \\
$\mathbf{g}_t$ & Vec3 & Target waypoint & In bounds \\
$\psi_t$ & Scalar & Target yaw & $[-\pi,\pi]$ \\
$v_t$ & Scalar & Target speed & $[0,v_{\max}]$ \\
$h_t$ & Scalar & Target altitude & $[h_{\min},h_{\max}]$ \\
$r_t$ & Scalar & Acceptance radius & $[0,r_{\max}]$ \\
$\hat{s}_t$ & Int & Active step & $1\le \hat{s}_t \le L$ \\
$\gamma_t$ & Scalar & Confidence & $[0,1]$ \\
\bottomrule
\end{tabular}
\end{table}

\vspace{0.1cm}
\noindent \textbf{Control Layer.} The control layer executes $D_t$ at a higher frequency than the semantic cycle and realizes SLM-issued directives via reinforcement-learning (RL) policies.
Given waypoints or sub-mission commands, the RL controller maps them to autopilot-compatible setpoints.
Operating at a higher rate enables rapid compensation for sensing noise and newly emerging obstacles, maintaining flight stability.
Even when high-level decisions remain unchanged for multiple cycles, the RL layer continues closed-loop safety control, preserving safe and smooth motion under intermittent semantic updates.

\subsection{Dynamic Role Orchestration}
\label{sec:role_orch}
\begin{figure}[h!]
\centering
\includegraphics[width=1\linewidth]{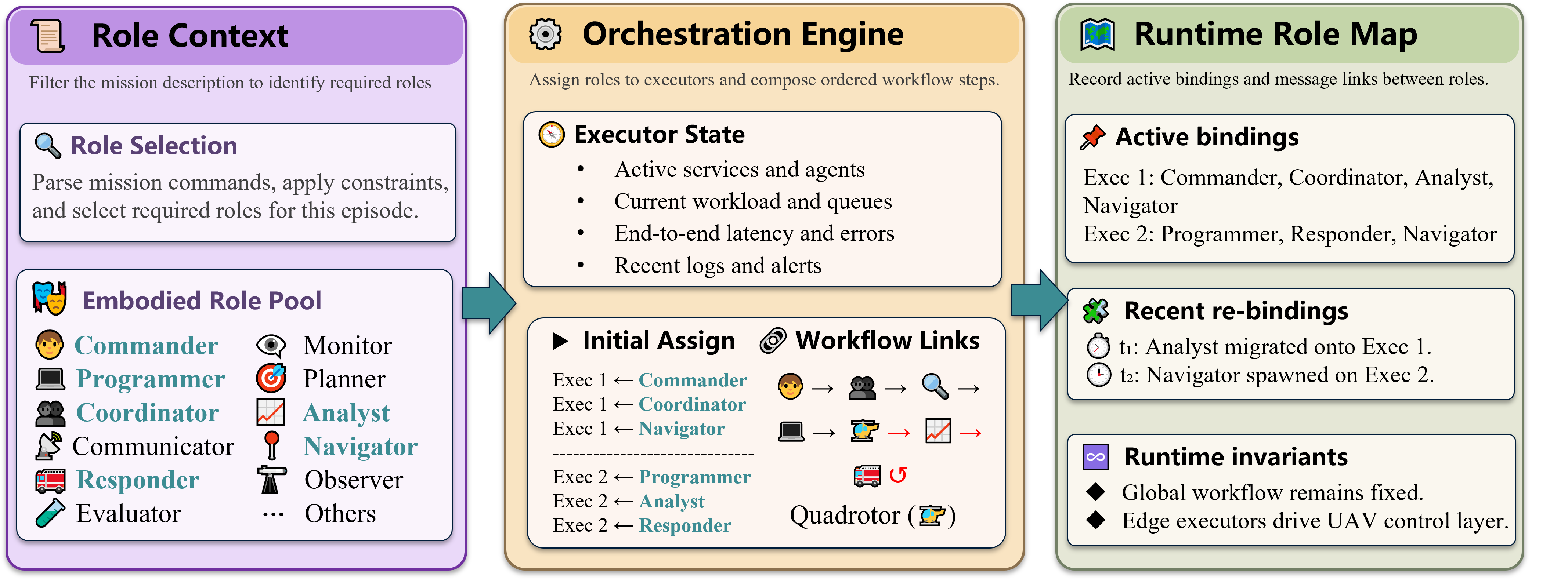}
\caption{\textbf{Dynamic role orchestration in AERIS.} An offline LLM-generated role-based workflow and the embodied role pool provide semantic context for the orchestration engine, which maintains a runtime role map to update role–executor bindings.}
\label{fig:roles}
\end{figure}

Fig.~\ref{fig:roles} illustrates how these role abstractions are used across three stages: offline role selection, edge orchestration, and runtime adaptation. Before deployment, AERIS organizes the models already available at the edge into an executable role pool $\mathcal{R}=\{r_i\}$. The role context module filters task descriptions and related constraints, selects the roles compatible with the current mission, and extracts the task-specific requirements that guide subsequent orchestration.\par

When a mission begins, the orchestration engine receives the pre-constructed role context and initializes a workflow based on the selected roles.
Specifically, AERIS generates the workflow through a single offline call to the LLM, thereby defining the execution sequence of roles and the information flow paths. Based on the initial measurement results of the established workflow and the status of the executors, the orchestration engine performs the initial binding operation: selecting appropriate nodes from the executor set $\mathcal{E} = \{e_j\}$ to match corresponding hosting executors for each role in the role set $\mathcal{R}$, thereby obtaining the initial runtime role mapping $\pi_0: \mathcal{R} \rightarrow \mathcal{E}$.
During the task execution process, the runtime mapping will clearly identify the hosting executor corresponding to each role instance. As the load and latency conditions fluctuate dynamically, the orchestration engine will update $\pi_t$, only re-binding the minimum number of roles, and migrating the running processes of these roles among the edge executors, while maintaining the stability of the workflow structure unchanged.
This dynamic re-binding method can not only completely preserve the original structure of the workflow, but also effectively enhance the robustness of the downstream closed-loop collaboration, ensuring the overall operational stability.
\subsection{Communication Hub}
\label{sec:comm_hub}
The Communication Hub is the runtime middleware that bridges high-level instruction reasoning and low-level execution.
It provides schema-aware message passing and implements an attention--subgoal alignment mechanism (ATT) to keep communications aligned with the currently active instruction step.
Instead of treating attention as an internal feature of a reasoning model, we realize ATT in the hub as an arbitration operator $\eta$ that annotates each outgoing message with a compact focus context.

\begin{figure}[h]
    \centering
    \includegraphics[width=0.95\linewidth]{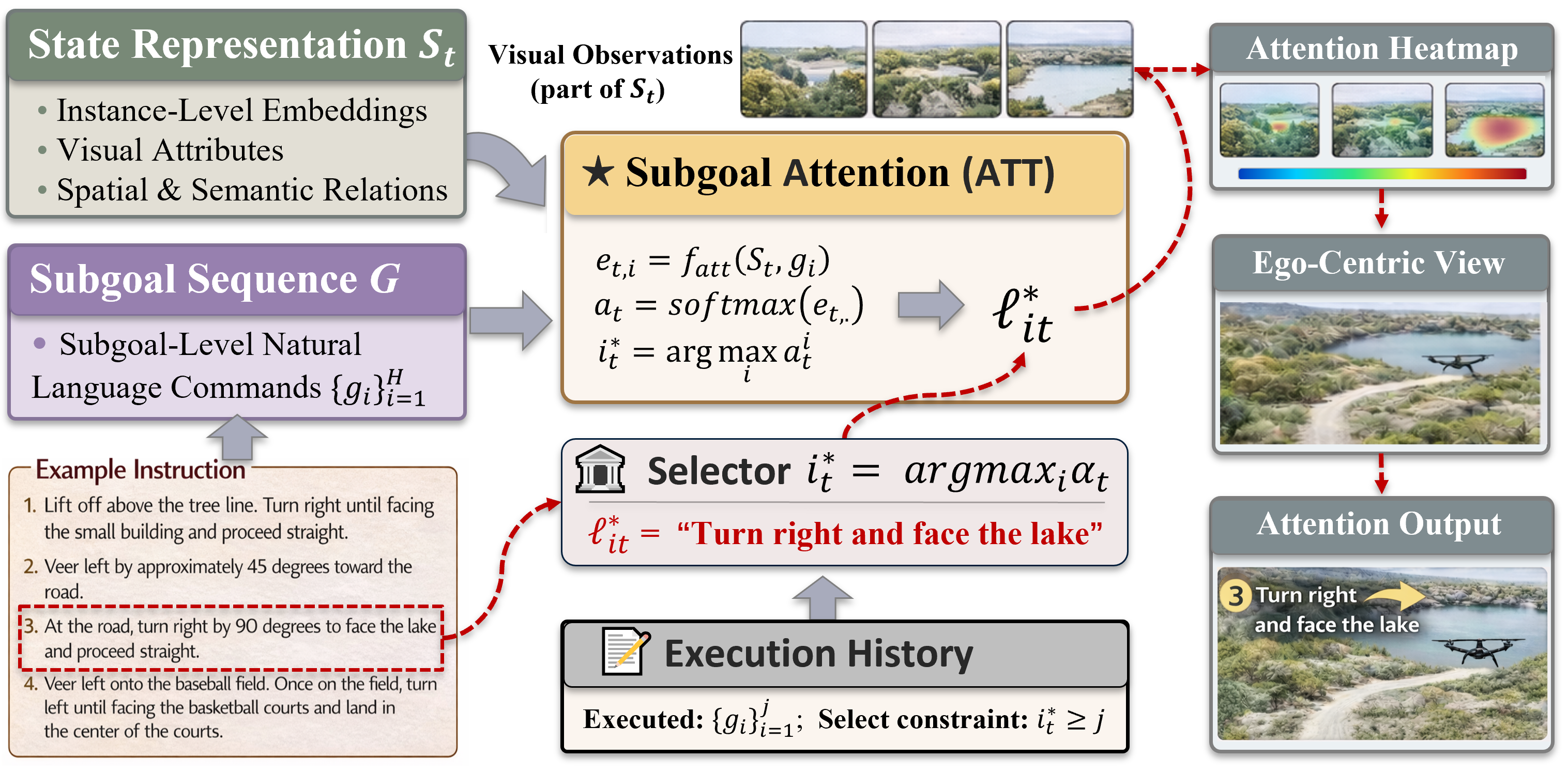}
    \caption{\textbf{Illustration of the ATT mechanism in the Communication Hub.} ATT computes attention weights $\alpha_t$ over the segmented instruction subgoals based on the agent's state $S_t$, identifies the most relevant subgoal, and annotates outgoing messages with this context. }
    \label{fig:att}
\end{figure}
\vspace{0.1cm}
\noindent \textbf{ATT Mechanism.} As illustrated in Fig.~\ref{fig:att}, given the current state $S_t$ and a set of $H$ instruction subgoals $\mathcal{G}=\{g_i\}_{i=1}^{H}$, the hub computes subgoal relevance scores
\begin{equation}
  e_{t,i} = f_{\text{att}}(S_t, g_i),
\end{equation}
where $f_{\text{att}}$ is a lightweight compatibility function.
These relevance scores $\{e_{t,i}\}_{i=1}^{H}$ are normalized into an attention distribution
$\alpha_t=(\alpha_{t,1},\ldots,\alpha_{t,H})$ via
\begin{equation}
\alpha_{t,i} = \frac{\exp(e_{t,i})}{\sum_{j=1}^H \exp(e_{t,j})}, \qquad i = 1,\ldots,H.
\end{equation}
and the active subgoal index is selected as
\begin{equation}
i^*_t = \arg\max_i \alpha_{t,i}.
\end{equation}
The arbitration operator $\eta$ annotates each outgoing message with $i_t^*$ and can optionally prioritize latency-critical messages that are most relevant to the active step.
This yields an interpretable attention trace and provides a direct signal for stepwise instruction-following evaluation (e.g., ASR in Sec.~\ref{sec:exp_vln_main}).

\vspace{0.1cm}
\noindent \textbf{Validation, Binding, and Additional Hub Services.}
Beyond ATT-based arbitration, the Communication Hub also serves as the runtime boundary between semantic reasoning and low-level execution. Given the schema-constrained decision $D_t$ produced by the semantic layer, the hub first checks whether it satisfies the decision schema in Table~\ref{tab:decision_schema}. Let $\mathcal{S}_D$ denote the set of schema-valid decisions. The validation rule is
\begin{equation}
\label{eq:decision_valid}
\mathrm{Valid}(D_t)=\mathbb{I}\!\left[D_t\in\mathcal{S}_D\right],
\end{equation}
where required-field presence, data types, and numerical ranges are checked deterministically. When available, the hub further applies conservative feasibility checks derived from the typed state $S_t$, such as occupancy-aware rejection or clamping of unsafe waypoint, speed, and altitude fields.

A valid decision is then bound to executable control targets, including waypoint, yaw, speed, altitude, and controller mode. If $D_t$ is invalid or the semantic cycle misses its heartbeat deadline, the hub rejects the candidate output and reuses the last valid decision, thereby preventing malformed semantic outputs from directly affecting the high-rate controller. In addition, the hub maintains topic-based publish/subscribe routing, ordered message delivery, an executor resource registry, telemetry filtering, and code-execution utilities for grounding high-level decisions into low-level actuator commands.

\subsection{Benchmarks}
\label{sec:benchmarks}
\noindent \textbf{Environment and observations.} We adopt the AerialVLN benchmark~\cite{liu2023aerialvln}, a city-scale UAV VLN task built on Unreal Engine.
Each VLN episode is defined by a natural-language instruction $l$ paired with a reference trajectory.
The agent receives egocentric RGB observations and must navigate by grounding language to visual landmarks without access to privileged maps or goal coordinates.

\begin{figure}[h]
  \centering
  \includegraphics[width=0.98\linewidth]{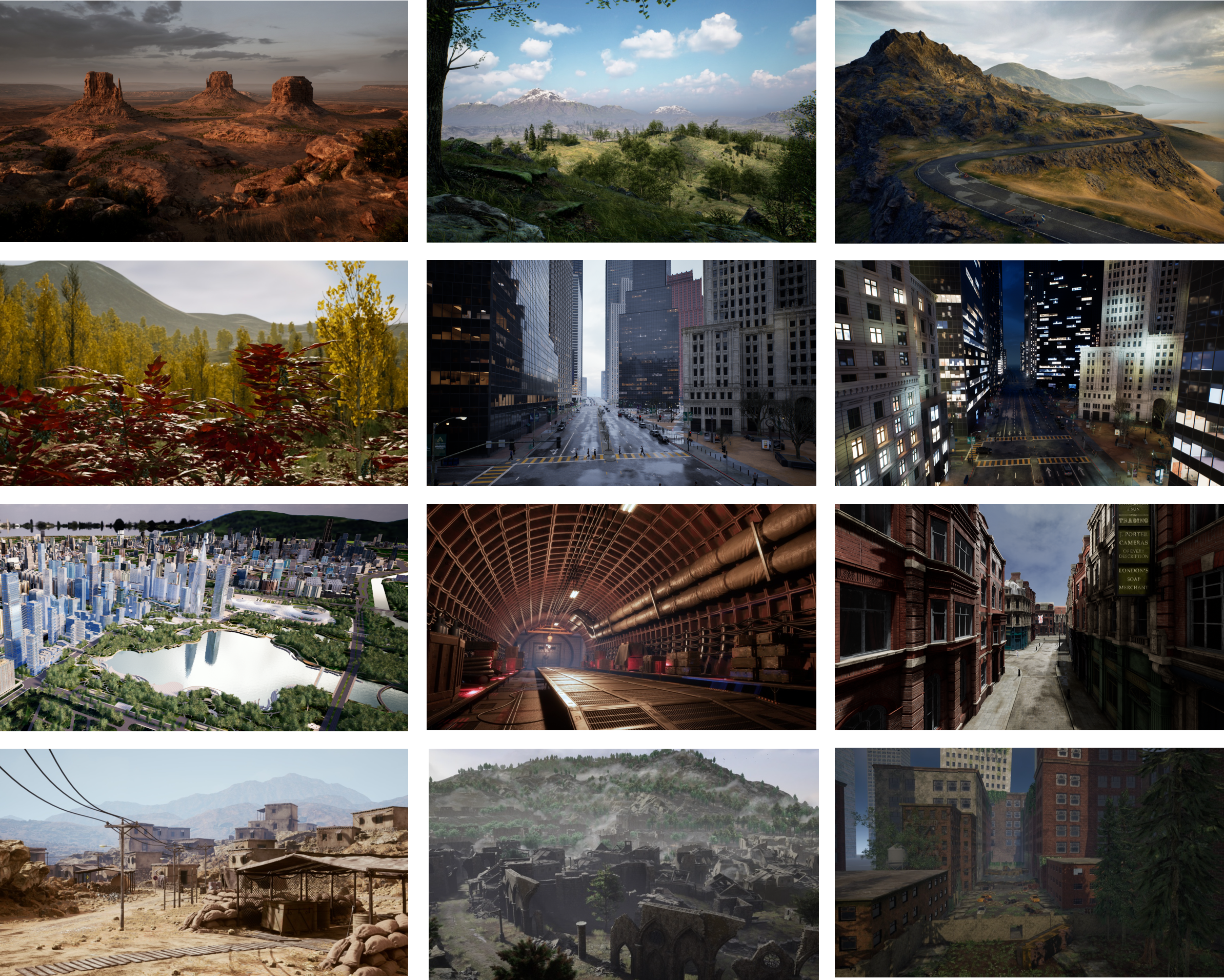}
  \caption{\textbf{Representative scenes in the VLN benchmark.}
  Beyond the original urban environments, we add additional open-area scenes,
  resulting in diverse layouts spanning both dense city structures and wide-open spaces under varied lighting conditions.}
  \label{fig:vln_env}
\end{figure}
\vspace{0.1cm}
\noindent \textbf{Dataset splits and augmentation.}
We follow the standard AerialVLN protocol and use its official {Train}, {Validation-Unseen (VU)}, and {Test-Unseen (TU)} splits.
Models are trained on {Train}, hyperparameters are tuned on {VU}, and results are reported on both {VU} and {TU}.
To broaden environmental diversity beyond the official scene collection, we additionally introduce a set of Unreal Engine scenes.
As illustrated in Fig.~\ref{fig:vln_env}, the added scenes span diverse open-area terrains (e.g., forests and deserts), complementing the original dense urban layouts and providing more distinctive landmarks.
This expansion reduces reliance on a single urban style and enables a stricter evaluation of landmark-driven, stepwise ATT grounding in unseen environments.
For difficulty analysis, we treat the original urban scenes as {Dense Airspace} and the newly added scenes as {Open Airspace}.\par

\begin{table*}[t]
\centering
\footnotesize
\setlength{\tabcolsep}{2.6pt}
\renewcommand{\arraystretch}{1.05}
\caption{VLN performance comparison on VU and TUwith breakdowns by Full/Open/Dense airspaces. }
\label{tab:vln_main}

\makebox[\textwidth][c]{%
\begin{tabular}{@{}lc
S[table-format=3.1] S[table-format=3.1] S[table-format=3.1]
S[table-format=2.1] S[table-format=2.1] S[table-format=2.1]
S[table-format=2.1] S[table-format=2.1] S[table-format=2.1]
S[table-format=2.1] S[table-format=2.1] S[table-format=2.1]
S[table-format=2.1] S[table-format=2.1] S[table-format=2.1]
S[table-format=2.1] S[table-format=2.1] S[table-format=2.1]
@{}}
\toprule
\multirow{2}{*}{Method} & \multirow{2}{*}{Split}
& \multicolumn{3}{c}{NE/m$\downarrow$}
& \multicolumn{3}{c}{SR/\%$\uparrow$}
& \multicolumn{3}{c}{OSR/\%$\uparrow$}
& \multicolumn{3}{c}{nDTW/\%$\uparrow$}
& \multicolumn{3}{c}{SPL/\%$\uparrow$}
& \multicolumn{3}{c}{ASR/\%$\uparrow$} \\
\cmidrule(lr){3-5}\cmidrule(lr){6-8}\cmidrule(lr){9-11}\cmidrule(lr){12-14}\cmidrule(lr){15-17}\cmidrule(lr){18-20}
& &
\multicolumn{1}{c}{Full} & \multicolumn{1}{c}{Open} & \multicolumn{1}{c}{Dense}
& \multicolumn{1}{c}{Full} & \multicolumn{1}{c}{Open} & \multicolumn{1}{c}{Dense}
& \multicolumn{1}{c}{Full} & \multicolumn{1}{c}{Open} & \multicolumn{1}{c}{Dense}
& \multicolumn{1}{c}{Full} & \multicolumn{1}{c}{Open} & \multicolumn{1}{c}{Dense}
& \multicolumn{1}{c}{Full} & \multicolumn{1}{c}{Open} & \multicolumn{1}{c}{Dense}
& \multicolumn{1}{c}{Full} & \multicolumn{1}{c}{Open} & \multicolumn{1}{c}{Dense} \\
\midrule

Random  & VU
& 149.7 & 136.8 & 167.2
& 0.0   & 0.0   & 0.0
& 0.0   & 0.0   & 0.0
& 0.0   & 0.0   & 0.0
& 0.0   & 0.0   & 0.0
& \multicolumn{1}{c}{\textit{--}} & \multicolumn{1}{c}{\textit{--}} & \multicolumn{1}{c}{\textit{--}} \\

Seq2Seq & VU
& 218.9 & 200.0 & 244.5
& 2.3   & 2.7   & 1.8
& 11.7  & 13.6  & 9.1
& 30.4  & 35.1  & 24.0
& 1.0   & 1.2   & 0.7
& \multicolumn{1}{c}{\textit{--}} & \multicolumn{1}{c}{\textit{--}} & \multicolumn{1}{c}{\textit{--}} \\

CMA     & VU
& 172.1 & 157.3 & 192.3
& 3.2   & 3.7   & 2.5
& 16.0  & 18.6  & 12.4
& 34.4  & 39.7  & 27.2
& 1.5   & 1.8   & 1.1
& \multicolumn{1}{c}{\textit{--}} & \multicolumn{1}{c}{\textit{--}} & \multicolumn{1}{c}{\textit{--}} \\

LAG     & VU
& 127.9 & 116.9 & 142.9
& 5.1   & 5.9   & 4.0
& 10.5  & 12.2  & 8.2
& 27.5  & 31.7  & 21.8
& 2.7   & 3.3   & 1.9
& \multicolumn{1}{c}{\textit{--}} & \multicolumn{1}{c}{\textit{--}} & \multicolumn{1}{c}{\textit{--}} \\

Human   & VU
& 19.1 & 17.9 & 20.7
& 81.7 & 85.5 & 76.5
& 84.7 & 89.4 & 78.4
& 95.5 & 95.3 & 95.8
& 69.9 & 75.2 & 62.7
& 92.2 & 93.6 & 90.3 \\

\rowcolor{blue!10}
Ours    & VU
& 79.4 & 72.6 & 88.7
& 25.8 & 29.8 & 20.4
& 44.3 & 51.6 & 34.4
& 64.9 & 74.9 & 51.3
& 15.0 & 18.5 & 10.2
& 83.1 & 88.4 & 76.0 \\

\addlinespace
\midrule

Random  & TU
& 148.5 & 136.2 & 165.2
& 0.0   & 0.0   & 0.0
& 0.0   & 0.0   & 0.0
& 0.0   & 0.0   & 0.0
& 0.0   & 0.0   & 0.0
& \multicolumn{1}{c}{\textit{--}} & \multicolumn{1}{c}{\textit{--}} & \multicolumn{1}{c}{\textit{--}} \\

Seq2Seq & TU
& 214.6 & 196.9 & 238.7
& 2.2   & 2.4   & 1.9
& 9.4   & 10.3  & 8.1
& 31.8  & 34.9  & 27.6
& 0.9   & 1.1   & 0.7
& \multicolumn{1}{c}{\textit{--}} & \multicolumn{1}{c}{\textit{--}} & \multicolumn{1}{c}{\textit{--}} \\

CMA     & TU
& 178.5 & 163.8 & 198.5
& 3.9   & 4.3   & 3.4
& 13.1  & 14.4  & 11.4
& 35.9  & 39.4  & 31.1
& 1.9   & 2.2   & 1.5
& \multicolumn{1}{c}{\textit{--}} & \multicolumn{1}{c}{\textit{--}} & \multicolumn{1}{c}{\textit{--}} \\

LAG     & TU
& 128.3 & 117.7 & 142.7
& 4.5   & 4.9   & 3.9
& 11.6  & 12.7  & 10.1
& 28.9  & 31.7  & 25.1
& 2.3   & 2.6   & 1.8
& \multicolumn{1}{c}{\textit{--}} & \multicolumn{1}{c}{\textit{--}} & \multicolumn{1}{c}{\textit{--}} \\

\rowcolor{blue!10}
Ours    & TU
& 85.4 & 78.3 & 95.0
& 20.5 & 22.5 & 17.7
& 31.8 & 34.9 & 27.6
& 66.0 & 72.5 & 57.2
& 11.4 & 13.5 & 8.5
& 73.6 & 78.5 & 67.0 \\
\bottomrule
\end{tabular}%
}
\end{table*}

\section{Experiments}
\label{sec:experiments}

\subsection{Experimental Setup}
\noindent\textbf{Implementations.}
All simulation experiments were carried out in accordance with the VLN evaluation scheme described.
AERIS employs a fixed heartbeat scheduling mechanism: the semantic stack generates decisions $D_t$ that comply with the schema constraints every 2 seconds, while the 50 HZ controller tracks the instructions in real time.
If a cycle misses the deadline, the system will automatically reuse the previous decision, thereby decoupling the high-level delay from the low-level flight safety.

\noindent\textbf{Baselines.} We compare against: \textit{(i)} Random, a non-learning policy that uniformly samples feasible actions from the same action interface;
\textit{(ii)} {Seq2Seq}~\cite{anderson2018}, an instruction-conditioned encoder--decoder trained for stepwise action prediction;
\textit{(iii)} {CMA}~\cite{krantz2020beyond}, a cross-modal attention agent that fuses language and visual features for navigation decisions;
and \textit{(iv)} {LAG}~\cite{liu2023aerialvln}, which uses look-ahead signals to improve long-horizon action selection.
All baselines are trained and evaluated with the same observation interfaces and stopping rule as AERIS.\par

\noindent\textbf{Evaluation metrics.} Following the standard VLN evaluation protocol~\cite{krantz2020beyond}, we adopt Success Rate (SR), Oracle Success Rate (OSR), and Navigation Error (NE).
SR counts the proportion of all test episodes where the final stopping position is within a success radius $\tau{=}20$\ m of the target.
OSR measures the proportion of episodes that enter the $20$m-neighborhood of the target at any point during execution.
NE is the Euclidean distance between the final position and the goal position.
To go beyond terminal-only evaluation of "only evaluating terminal accuracy", we additionally introduce the Normalized Dynamic Time Warping (nDTW)~\cite{ilharco2019dtw} to measure similarity between the executed and the reference trajectory.
We also report Success weighted by Path Length (SPL)~\cite{anderson2018evaluation}, which evaluates navigation efficiency conditioned on success:
\begin{equation}
\mathrm{SPL}=\frac{1}{N}\sum_{i=1}^{N} S_i \cdot \frac{l_i}{\max(p_i,l_i)}\,,
\end{equation}
where $S_i\in\{0,1\}$ indicates success under radius $\tau$, $l_i$ is the shortest path distance from the start point to the target, and $p_i$ is the actual path length.
Finally, we introduce the {Attention-weighted Success Rate (ASR)}, which judges the completion of subgoals through the attention trace:

\begin{equation}
\mathrm{ASR} =  \frac{1}{N}\sum_{i=1}^{N}\frac{c_i}{K_i},
\end{equation}
where $K_i$ is the number of subgoals in episode $i$, and $c_i \in [0,K_i]$ is the number of completed subgoals inferred from the attention trace (ranging from 0 to \( K_i \)).
For methods that do not produce attention traces, ASR is not applicable and is denoted as ``--''.
\par

\subsection{Main Results}
\label{sec:exp_vln_main}
Table~\ref{tab:vln_main} reports VLN results of AERIS against standard baselines and human pilots on VU and TU splits, with breakdowns by full, open and dense airspace.

\noindent\textbf{Overall Performance.}
Across all splits and airspace densities, AERIS outperforms all baselines, with the largest gains in SR and OSR, demonstrating more reliable long-horizon instruction following and goal reaching in unseen environments.
All methods degrade in {Dense Airspace} due to increased clutter, occlusions, and weaker long-range landmark cues; however, AERIS preserves a clear advantage, indicating stronger robustness of the closed-loop stack under challenging perceptual and interaction conditions.

\noindent\textbf{Path quality (NE, nDTW).}
In addition to simply evaluating system performance based on task success or failure, AERIS achieves higher trajectory fidelity of the flight trajectory throughout the entire task execution process.
It can stably reduce the NE, which means that the final target positioning will be more accurate.
At the same time, it significantly improves nDTW, indicating that the trajectory alignment throughout the flight process has also been optimized.
This performance advantage is particularly evident in dense airspace where obstacle obstructions and environmental noise will cause the positioning deviation to increase.

\noindent\textbf{Path efficiency (SPL).} AERIS improves not only the goal-reaching ability but also path efficiency, achieving consistently higher SPL than all baselines.
This indicates that, when successful, trajectories are typically shorter and more direct rather than exhibiting prolonged detours.
The advantage persists across airspace densities: although SPL decreases for all methods in {Dense Airspace} due to obstacle avoidance and occlusions, AERIS maintains a clear margin.
\par

\noindent\textbf{Human upper bound (ASR vs.\ SR).} We treat human trajectories as an empirical upper bound, as they are reference demonstrations flown by human operators rather than generated by a deployable autonomous policy.
Compared with human pilots, AERIS is much closer in ASR than in SR, since ASR credits subgoal completion even in failed episodes while SR is purely binary.
The remaining gap is mainly due to limited online recovery after losing the active subgoal: when a landmark is missed or alignment drifts, AERIS often struggles to re-localize and re-anchor progress, whereas human pilots typically recover by actively exploring viewpoints and exploiting contextual cues.

\subsection{Additional studies}
\begin{figure}[h!]
  \centering
  \includegraphics[width=\linewidth]{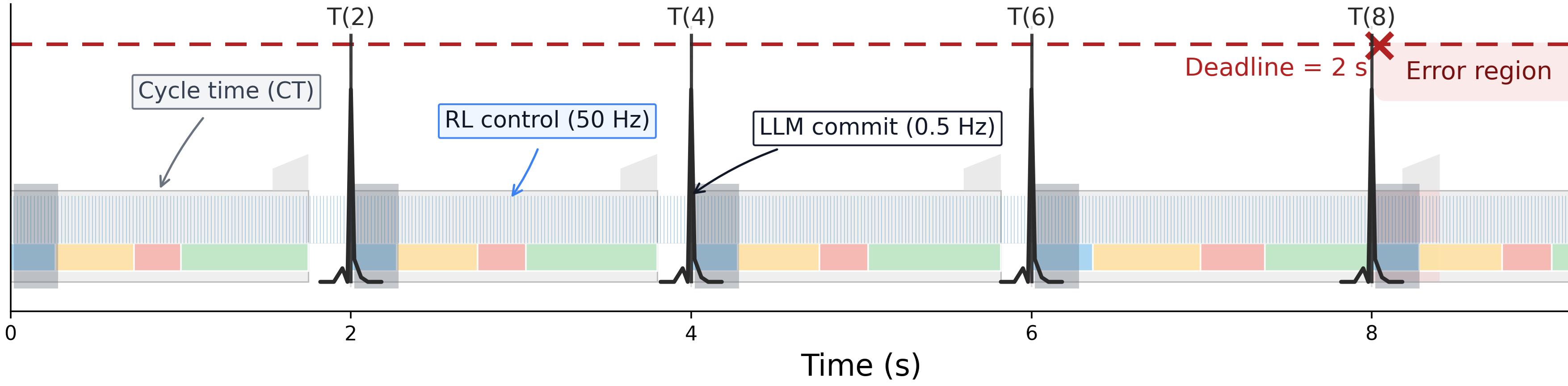}
\caption{\textbf{Heartbeat-timed timing trace.}
  Gray bars show CT with stage breakdown: Perception (blue), Reasoning (orange), Routing (pink), and Binding (green).
  Black spikes mark heartbeat commits; the 2\,s budget is used to judge overruns.}
  \label{fig:rt_trace}
\end{figure}

\noindent\textbf{Runtime cost and execution reliability.}
Fig.~\ref{fig:rt_trace} analyzes the runtime cost and closed-loop reliability of the AERIS system in the heartbeat timing mode. Here, we set the running frequency of the semantic stack to 0.5 HZ, which corresponds to a 2s heartbeat cycle budget. The black peaks in the figure represent the moments when instructions are submitted; simultaneously, the underlying controller runs at a frequency of 50 HZ.

During each heartbeat cycle, we recorded the cycle time (CT), which is the total delay from the start of the instruction to its submission, and calculated the execution error rate (EER). As shown in Table \ref{tab:rt_cost_eer}, the cycle time increases as the complexity of the scenario increases: compared to open airspace, the average cycle time in dense airspace is longer, making it more prone to timeouts and heartbeat synchronization errors. This impact directly reflects the increase in the execution error rate. Therefore, during the execution of long-duration navigation tasks, even a few invalid cycles can have a significant impact on the temporal stability.

\begin{table}[h!]
\centering
\small
\setlength{\tabcolsep}{6pt}
\renewcommand{\arraystretch}{1.10}
\caption{Timing and reliability under a 2\,s heartbeat.}
\label{tab:rt_cost_eer}
\begin{tabular}{lcc}
\toprule
\textbf{Split} & \textbf{CT (s) mean$\pm$std / P95} & \textbf{EER (\%)} \\
\midrule
Open Airspace  & 1.31$\pm$0.18 / 1.64 & 0.4 \\
Full           & 1.42$\pm$0.20 / 1.78 & 1.9 \\
Dense Airspace & 1.63$\pm$0.28 / 2.06 & 2.8 \\
\bottomrule
\end{tabular}
\end{table}
\par
\noindent\textbf{Ablation studies.}
We quantify the contribution of each core component via controlled ablations on the AerialVLN {Test-Unseen} split under consistent experimental configurations.
We consider four variants: {No-ATT} disables hub-side ATT mechanism; {Single-SLM} replaces the role-adaptive compact SLM stack with a single fixed SLM variant; {No-DS} eliminates dynamic re-binding scheme; and {Zero-shot} removes task-specific alignment and applies generic models.\par
Table~\ref{tab:ablation_vln_tu} indicates that the complete AERIS achieves the best performance in terms of navigation quality and runtime reliability. In contrast, all the variants show varying degrees of decline.
{No-ATT} causes the largest drop and the most severe timing failures, indicating that the sub-goal instruction focusing is critical for stable long-horizon progress.
{Zero-shot} substantially harms grounding and trajectory fidelity, supporting the necessity for task-specialized alignment to translate instructions into executable decisions.
{Single-SLM} yields a consistent but moderate degradation and higher EER, suggesting that the multi-scale stack better balances reasoning capacity and responsiveness.
{No-DS} primarily impacts timing stability with relatively smaller changes in navigation metrics, implying that dynamic scheduling mainly improves robustness under variable load.
Overall, the ablations indicate that AERIS requires (i) hub-side step alignment for instruction-consistent coordination, (ii) task-aligned compact reasoning for grounded decisions, and (iii) dynamic re-binding for timing robustness.
\begin{figure}[h!]
  \centering
  \includegraphics[width=\linewidth]{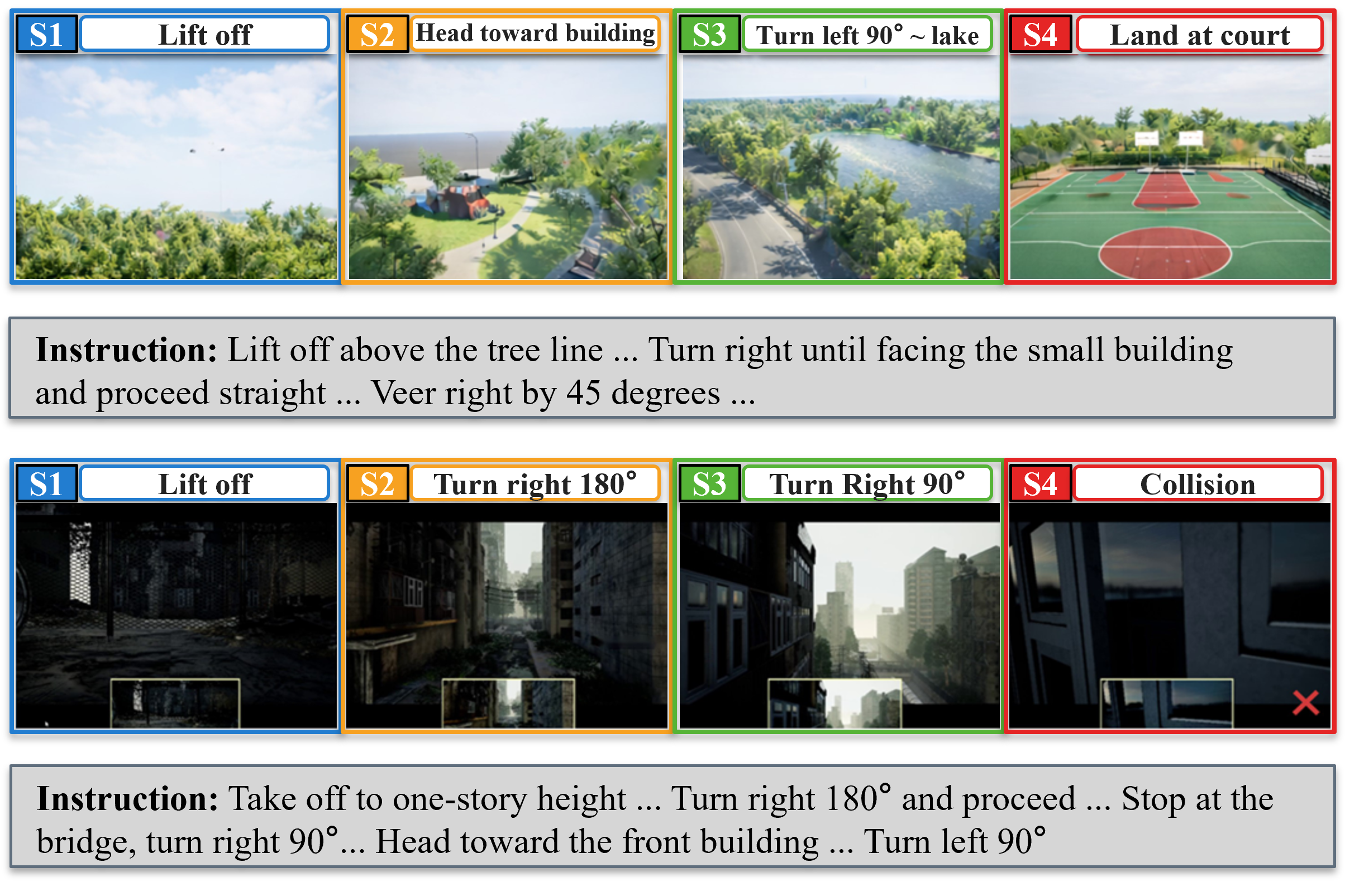}
  \caption{\textbf{Qualitative rollouts with AERIS.}
  Top: a successful instruction-following trajectory with representative subgoals.
  Bottom: a failure case where the UAV eventually collides, illustrating error accumulation under challenging visual conditions.}
  \label{fig:qual_cases}
\end{figure}

\begin{figure}[ht]
  \centering
  \includegraphics[width=1\linewidth]{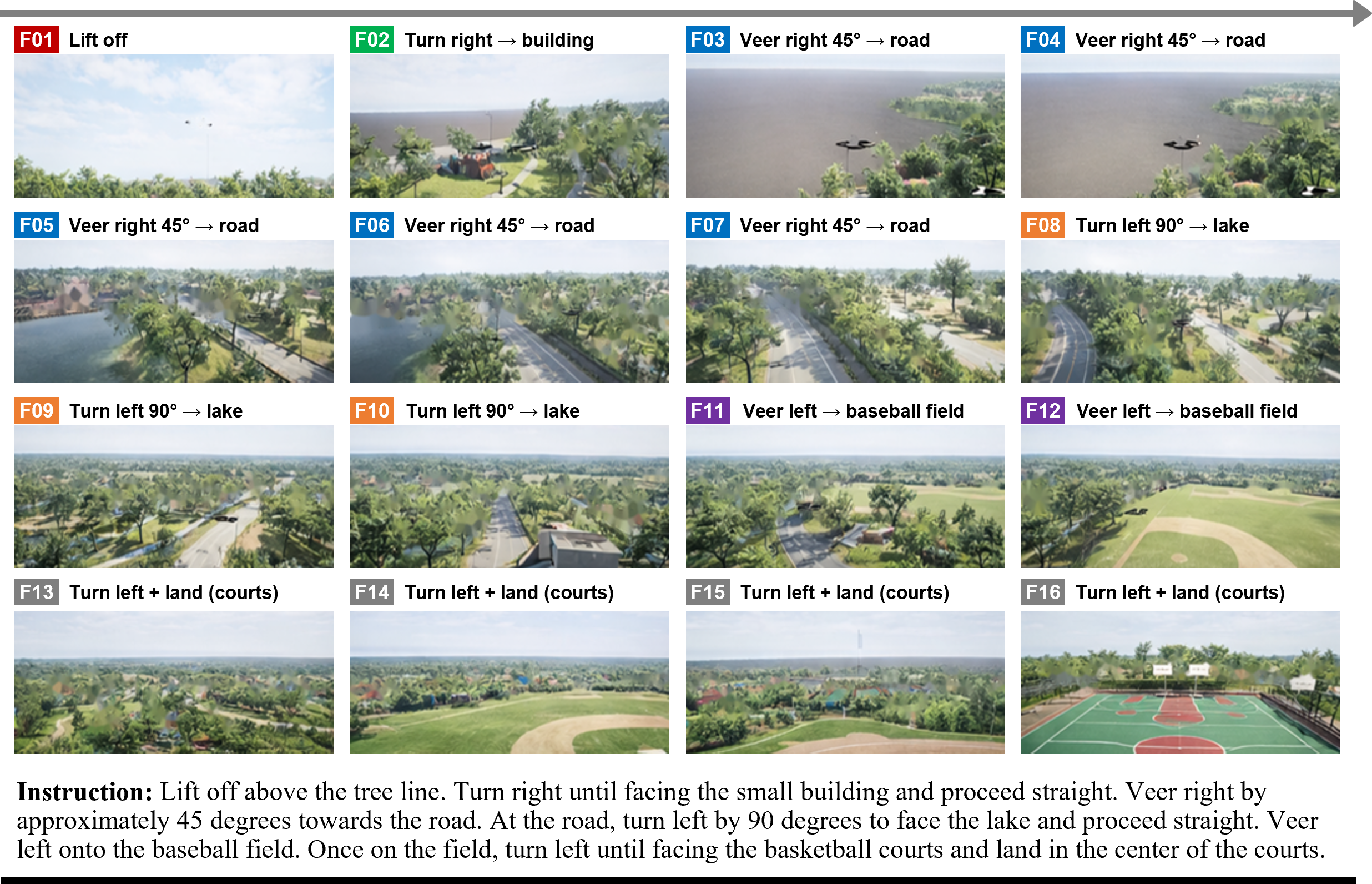}
  \caption{\textbf{VLN visualization in simulation.}
  Sampled first-person frames are annotated with the active instruction step emitted by the Communication Hub, illustrating stepwise grounding and progress over a long-horizon trajectory.}
  \label{fig:vis_vln_airsim}
\end{figure}

\begin{table}[h!]
  \centering
  \caption{{Ablation results on the VLN Test-Unseen split (Full).}}
  \label{tab:ablation_vln_tu}
  \setlength{\tabcolsep}{3.2pt}
  \renewcommand{\arraystretch}{1.05}
  \small
  \resizebox{\linewidth}{!}{
  \begin{tabular}{lcccccccc}
    \toprule
    \textbf{Method (Variant)} &
    \textbf{CT (s)}$\downarrow$ &
    \textbf{EER (\%)}$\downarrow$ &
    \textbf{NE}$\downarrow$ &
    \textbf{SR}$\uparrow$ &
    \textbf{OSR}$\uparrow$ &
    \textbf{nDTW}$\uparrow$ &
    \textbf{SPL}$\uparrow$ &
    \textbf{ASR}$\uparrow$ \\
    \midrule
    AERIS (Full)
      & 1.42$\pm$0.20 & 1.9
      & 85.4 & 20.5\% & 31.8\%
      & 66.0 & 11.4\% & 73.6\% \\
    \midrule
    \ \ -- No-ATT
      & 4.55$\pm$1.48 & 12.5
      & 120.0 & 5.0\%  & 10.8\%
      & 52.0 & 2.7\%  & -- \\
    \ \ -- Single-SLM
      & 1.65$\pm$0.32 & 4.8
      & 99.7 & 17.1\% & 28.4\%
      & 61.5 & 9.3\%  & 65.0\% \\
    \ \ -- No-DS
      & 1.60$\pm$0.34 & 5.1
      & 90.1 & 18.3\% & 30.3\%
      & 64.0 & 9.9\%  & 70.2\% \\
    \ \ -- Zero-shot
      & 2.35$\pm$0.32 & 7.4
      & 135.3 & 3.3\%  & 5.5\%
      & 33.0 & 1.2\%  & 35.4\% \\
    \bottomrule
  \end{tabular}}
\end{table}

\vspace{0.1cm}
\noindent\textbf{Visualizations of success and failure cases.}
Fig.~\ref{fig:qual_cases} presents qualitative examples of the AERIS system executing instructions, contrasting a successful task execution and a failed one. In the successful scenario, AERIS follows the instructions, completing a series of subtasks in an orderly manner: taking off, then approaching a landmark building, turning left to the lake, and finally landing in the court area.
Even as the viewing angle changes constantly during flight, the UAV's flight path remains smooth and consistent. This shows that by updating intermediate goals in real time, the system can steadily drive the task forward.
In the failed case, the UAV completes the initial steps smoothly, but after entering a narrow corridor with complex visual conditions, its flight path gradually drifts away from the intended route. As localization and control errors accumulate over time, the UAV eventually flies toward an obstacle and collides with it.

\vspace{0.1cm}
\noindent\textbf{Instruction-step visualization in simulation.}
Fig.~\ref{fig:vis_vln_airsim} complements the trajectory-level rollouts with first-person frames from a simulated VLN episode in Unreal Engine and AirSim. The sampled frames are annotated with the active instruction step emitted by the Communication Hub, showing the progression through takeoff, heading calibration, road following, turning toward the lake, entering the sports field, and final landing. This visualization provides process-level evidence that the typed perception state, schema-constrained decision, and hub-side step alignment remain synchronized throughout a long-horizon rollout.

\section{Real-World Validation}
\label{sec:exp_real}
To examine whether {AERIS} can sustain stable closed-loop autonomy in physical settings, we conduct two real-world experiments targeting complementary capabilities: (i) stepwise instruction execution over long-horizon tasks, and (ii) time-sensitive coordination for multi-UAV formation flight.

\begin{figure}[h]
  \centering
  \includegraphics[width=\linewidth]{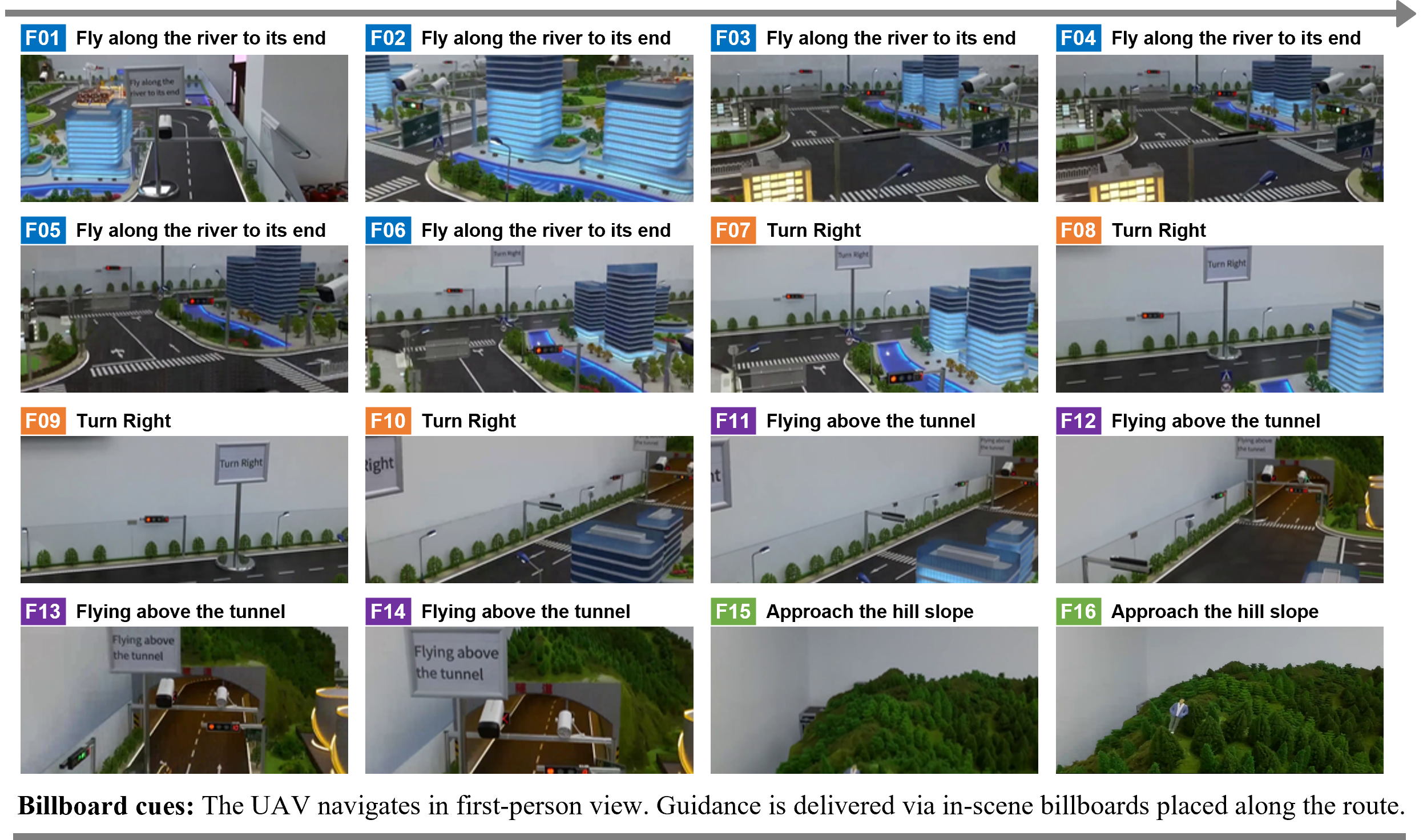}
  \caption{\textbf{Physical VLN rollout with in-scene cues.}
  A first-person sequence from the sandbox experiment guided by billboard cues placed along the route. The rollout illustrates step-triggered behaviors under explicit visual guidance while preserving schema-constrained, directly executable decisions.}
  \label{fig:vis_vln_cues}
\end{figure}

\vspace{0.1cm}
\noindent \textbf{Physical VLN in Sandbox Environment.}
To evaluate instruction alignment and dynamic planning for long-horizon, multi-step execution, we build a tabletop-scale controlled sandbox environment.
Instead of using bulky and costly physical landmarks, we deploy lightweight billboard-style signs as subgoal markers.
This design provides a step-alignment signal analogous to our attention--subgoal tracking while significantly reducing experimental cost.
During the experiment, {AERIS} runs on an on-site ground station, and the UAV streams real-time first-person-view (FPV) video to the station, enabling closed-loop perception, instruction-conditioned decision making, and command generation. Fig.~\ref{fig:vis_vln_cues} presents a representative first-person rollout from this sandbox deployment. Billboard cues placed along the route provide explicit visual guidance, and the sequence shows stage-triggered behaviors such as river following, turning, and flying above the tunnel while preserving the same typed interface and schema-constrained outputs used in simulation.

\begin{figure}[h]
  \centering
  \includegraphics[width=0.98\linewidth]{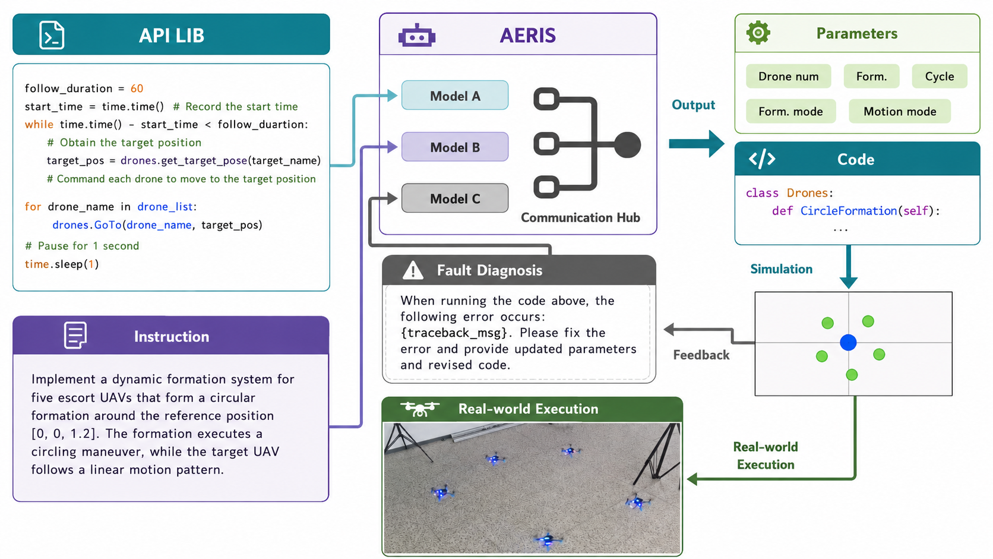}
  \caption{\textbf{Real-world formation pipeline of AERIS.}
  Given a natural-language instruction and an API-level code context, the Communication Hub produces schema-constrained formation parameters and executable code. The program is validated in simulation, deployed on physical UAVs, and iteratively refined via feedback-driven fault diagnosis when runtime errors occur.}
  \label{fig:formation_realworld}
\end{figure}
\begin{figure}[h]
  \centering
  \includegraphics[width=\linewidth]{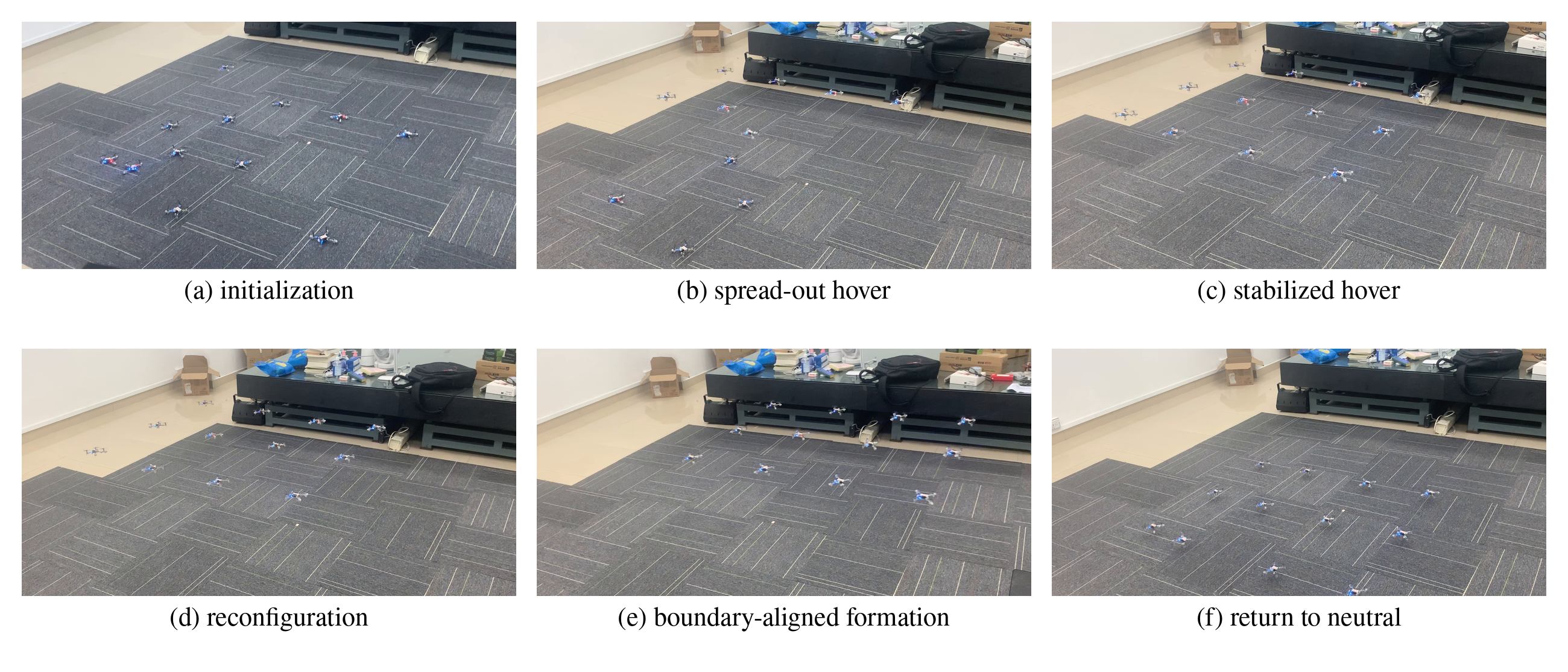}
  \caption{\textbf{Real-world formation sequence visualization.}
  Key frames sampled from an indoor multi-UAV flight illustrate the evolution of the team configuration over time, including initialization, spread-out hover, stabilized hover, reconfiguration, boundary-aligned formation, and return to neutral.}
  \label{fig:formation_sequence}
\end{figure}

\vspace{0.1cm}
\noindent \textbf{Multi-UAV Formation Flight Experiment.}
To validate real-time responsiveness and coordinated control, we deploy a swarm of micro-UAVs equipped with short-range wireless communication.
High-level orchestration is centrally executed on the on-site ground station, while each UAV independently runs a high-frequency onboard tracking controller to stabilize its own flight.
We issue formation tasks to the swarm via natural language and dynamically adjust the formation geometry on demand during maneuvering, testing both coordination latency and robustness under practical wireless and sensing constraints. Figs.~\ref{fig:formation_realworld} and \ref{fig:formation_sequence} illustrate the formation experiment at the workflow and execution levels, respectively. Fig.~\ref{fig:formation_realworld} shows the AERIS pipeline from language instruction and API context to code generation, simulation validation, and physical deployment, while Fig.~\ref{fig:formation_sequence} presents the indoor flight sequence from initialization to landing. Additional real-world footage is provided in the accompanying demo video.

\section{Conclusion}
This paper presented AERIS, a role-driven aerial-edge framework for language-model-assisted autonomy under real-time UAV constraints. Rather than treating the language model as a monolithic planner, AERIS decomposes perception, semantic decision making, communication, and control into runtime roles with typed interfaces. The edge-optimized model stack provides structured scene states and schema-constrained decisions; the Communication Hub validates and grounds these decisions through attention--subgoal alignment; and the dynamic orchestration mechanism re-binds roles across heterogeneous executors as resource conditions change while preserving a stable workflow. Together, these designs connect high-level instruction understanding with executable low-level control without relying on continuous cloud access.

Experiments on the AerialVLN benchmark show that AERIS improves long-horizon instruction following while maintaining bounded cycle time and low execution error under heartbeat-timed operation. The ablation results further indicate that system performance depends on the joint effect of task-aligned compact models, hub-side subgoal alignment, and runtime role re-binding, rather than on any single component alone. Real-world sandbox and multi-UAV experiments provide additional evidence that the role-driven abstraction can be deployed under practical sensing, communication, and control constraints.

Despite these improvements, the results also reveal the remaining limitations of AERIS. The system still falls short of human-level robustness when visual landmarks are missed, accumulated deviations cause subgoal drift, or dense environments weaken long-range visual cues. These failure cases suggest that a stable runtime pipeline alone is insufficient for robust long-horizon navigation. Future work will focus on explicit recovery and backtracking roles, stronger re-localization under occlusion, and more reliable subgoal tracking in dense or visually ambiguous environments.

\bibliographystyle{IEEEtran}
\bibliography{ref}

@inproceedings{liu2023aerialvln,
  author    = {Liu, Shubo and Zhang, Hongsheng and Qi, Yuankai and Wang, Peng and Zhang, Yanning and Wu, Qi},
  title     = {{AerialVLN}: Vision-and-Language Navigation for {UAVs}},
  booktitle = {Proceedings of the IEEE/CVF International Conference on Computer Vision},
  year      = {2023},
  pages     = {15384--15394},
  doi       = {10.1109/ICCV51070.2023.01411}
}

@inproceedings{li2023camel,
  author    = {Li, Guohao and Hammoud, Hasan Abed Al Kader and Itani, Hani and Khizbullin, Dmitrii and Ghanem, Bernard},
  title     = {{CAMEL}: Communicative Agents for ``Mind'' Exploration of Large Language Model Society},
  booktitle = {Advances in Neural Information Processing Systems},
  year      = {2023},
  volume    = {36},
  pages     = {51991--52008}
}

@misc{yolov8_docs,
  author       = {{Ultralytics}},
  title        = {{Ultralytics YOLOv8} Documentation},
  year         = {2023},
  howpublished = {\url{https://docs.ultralytics.com/models/yolov8/}},
  note         = {Accessed: 2026-05-12}
}

@inproceedings{rtdetr_cvpr24,
  author    = {Zhao, Yian and Lv, Wenyu and Xu, Shangliang and Wei, Jinman and Wang, Guanzhong and Dang, Qingqing and Liu, Yi and Chen, Jie},
  title     = {{DETRs} Beat {YOLOs} on Real-Time Object Detection},
  booktitle = {Proceedings of the IEEE/CVF Conference on Computer Vision and Pattern Recognition},
  year      = {2024},
  pages     = {16965--16974},
  doi       = {10.1109/CVPR52733.2024.01605}
}

@article{ppyoloe_arxiv,
  author  = {Xu, Shangliang and Wang, Xinxin and Lv, Wenyu and Chang, Qinyao and Cui, Cheng and Deng, Kaipeng and Wang, Guanzhong and Dang, Qingqing and Wei, Shengyu and Du, Yuning and Lai, Baohua},
  title   = {{PP-YOLOE}: An Evolved Version of {YOLO}},
  journal = {CoRR},
  volume  = {abs/2203.16250},
  year    = {2022},
  url     = {https://arxiv.org/abs/2203.16250},
  note    = {arXiv:2203.16250}
}

@inproceedings{occworld_eccv24,
  author    = {Zheng, Wenzhao and Chen, Weiliang and Huang, Yuanhui and Zhang, Borui and Duan, Yueqi and Lu, Jiwen},
  title     = {{OccWorld}: Learning a {3D} Occupancy World Model for Autonomous Driving},
  booktitle = {Proceedings of the European Conference on Computer Vision},
  year      = {2024},
  pages     = {55--72},
  doi       = {10.1007/978-3-031-72624-8_4}
}

@inproceedings{nemo_eccv24,
  author    = {Huang, Zanming and Zhang, Jimuyang and Ohn-Bar, Eshed},
  title     = {Neural Volumetric World Models for Autonomous Driving},
  booktitle = {Proceedings of the European Conference on Computer Vision},
  year      = {2024},
  pages     = {195--213},
  doi       = {10.1007/978-3-031-72943-0_12}
}

@inproceedings{occ_vo_2024,
  author    = {Li, Heng and Duan, Yifan and Zhang, Xinran and Liu, Haiyi and Ji, Jianmin and Zhang, Yanyong},
  title     = {{OCC-VO}: Dense Mapping via {3D} Occupancy-Based Visual Odometry for Autonomous Driving},
  booktitle = {Proceedings of the IEEE International Conference on Robotics and Automation},
  year      = {2024},
  pages     = {17961--17967},
  doi       = {10.1109/ICRA57147.2024.10611516}
}

@article{jia2024tinyllavafactory,
  author  = {Jia, Junlong and Hu, Ying and Weng, Xi and Shi, Yiming and Li, Miao and Zhang, Xingjian and Zhou, Baichuan and Liu, Ziyu and Luo, Jie and Huang, Lei and Wu, Ji},
  title   = {{TinyLLaVA Factory}: A Modularized Codebase for Small-Scale Large Multimodal Models},
  journal = {CoRR},
  volume  = {abs/2405.11788},
  year    = {2024},
  doi     = {10.48550/arXiv.2405.11788},
  url     = {https://arxiv.org/abs/2405.11788}
}

@article{wang2024openuav,
  author  = {Wang, Xiangyu and Yang, Donglin and Wang, Ziqin and Kwan, Hohin and Chen, Jinyu and Wu, Wenjun and Li, Hongsheng and Liao, Yue and Liu, Si},
  title   = {Towards Realistic {UAV} Vision-Language Navigation: Platform, Benchmark, and Methodology},
  journal = {CoRR},
  volume  = {abs/2410.07087},
  year    = {2024},
  url     = {https://arxiv.org/abs/2410.07087},
  note    = {arXiv:2410.07087}
}

@article{krantz2020beyond,
  author  = {Krantz, Jacob and Wijmans, Erik and Majumdar, Arjun and Batra, Dhruv and Lee, Stefan},
  title   = {Beyond the Nav-Graph: Vision-and-Language Navigation in Continuous Environments},
  journal = {CoRR},
  volume  = {abs/2004.02857},
  year    = {2020},
  url     = {https://arxiv.org/abs/2004.02857},
  note    = {arXiv:2004.02857}
}

@inproceedings{anderson2018,
  author    = {Anderson, Peter and Wu, Qi and Teney, Damien and Bruce, Jake and Johnson, Mark and S{\"u}nderhauf, Niko and Reid, Ian and Gould, Stephen and van den Hengel, Anton},
  title     = {Vision-and-Language Navigation: Interpreting Visually-Grounded Navigation Instructions in Real Environments},
  booktitle = {Proceedings of the IEEE Conference on Computer Vision and Pattern Recognition},
  year      = {2018},
  pages     = {3674--3683},
  doi       = {10.1109/CVPR.2018.00387}
}

@article{anderson2018evaluation,
  author  = {Anderson, Peter and Chang, Angel X. and Chaplot, Devendra Singh and Dosovitskiy, Alexey and Gupta, Saurabh and Koltun, Vladlen and Kosecka, Jana and Malik, Jitendra and Mottaghi, Roozbeh and Savva, Manolis and Zamir, Amir R.},
  title   = {On Evaluation of Embodied Navigation Agents},
  journal = {CoRR},
  volume  = {abs/1807.06757},
  year    = {2018},
  url     = {https://arxiv.org/abs/1807.06757},
  note    = {arXiv:1807.06757}
}

@article{lou2026agentstrainer,
  author  = {Lou, Jiabin and Shi, Rongye and Wang, Haopeng and Yu, Ming-Ming and Wang, Yuanshuai and Wang, Qunbo and Wu, Wenjun},
  title   = {Agents Trainer: Automatically Training Multi-Agent Reinforcement Learning Models for Drone Swarm Using Language Model-Based Agents},
  journal = {IEEE Transactions on Automation Science and Engineering},
  year    = {2026},
  doi     = {10.1109/TASE.2026.3689345},
  note    = {Early access/forthcoming; verify final volume, issue, and pages before final proof}
}

@article{xiao2024clipvg,
  author  = {Xiao, Linhui and Yang, Xiaoshan and Peng, Fang and Yan, Ming and Wang, Yaowei and Xu, Changsheng},
  title   = {{CLIP-VG}: Self-Paced Curriculum Adapting of {CLIP} for Visual Grounding},
  journal = {IEEE Transactions on Multimedia},
  year    = {2024},
  volume  = {26},
  pages   = {4334--4347},
  doi     = {10.1109/TMM.2023.3321501}
}

@inproceedings{hu2022lora,
  author    = {Hu, Edward J. and Shen, Yelong and Wallis, Phillip and Allen-Zhu, Zeyuan and Li, Yuanzhi and Wang, Shean and Wang, Lu and Chen, Weizhu},
  title     = {{LoRA}: Low-Rank Adaptation of Large Language Models},
  booktitle = {Proceedings of the International Conference on Learning Representations},
  year      = {2022},
  url       = {https://openreview.net/forum?id=nZeVKeeFYf9}
}

@inproceedings{dettmers2023qlora,
  author    = {Dettmers, Tim and Pagnoni, Artidoro and Holtzman, Ari and Zettlemoyer, Luke},
  title     = {{QLoRA}: Efficient Finetuning of Quantized {LLMs}},
  booktitle = {Advances in Neural Information Processing Systems},
  year      = {2023},
  volume    = {36},
  pages     = {10088--10115}
}

@inproceedings{ilharco2019dtw,
  author    = {Magalhaes, Gabriel Ilharco and Jain, Vihan and Ku, Alexander and Ie, Eugene and Baldridge, Jason},
  title     = {General Evaluation for Instruction Conditioned Navigation Using Dynamic Time Warping},
  booktitle = {NeurIPS Workshop on Visually Grounded Interaction and Language},
  year      = {2019},
  url       = {https://arxiv.org/abs/1907.05446}
}

@inproceedings{ghosh2024octo,
  author    = {{Octo Model Team} and Ghosh, Dibya and Walke, Homer and Pertsch, Karl and others},
  title     = {{Octo}: An Open-Source Generalist Robot Policy},
  booktitle = {Proceedings of Robotics: Science and Systems},
  year      = {2024},
  doi       = {10.15607/RSS.2024.XX.090}
}

@inproceedings{khazatsky2024droid,
  author    = {Khazatsky, Alexander and Pertsch, Karl and Nair, Suraj and Balakrishna, Ashwin and others},
  title     = {{DROID}: A Large-Scale In-the-Wild Robot Manipulation Dataset},
  booktitle = {Proceedings of Robotics: Science and Systems},
  year      = {2024},
  doi       = {10.15607/RSS.2024.XX.120}
}

@inproceedings{yue2024deervla,
  author    = {Yue, Yang and Wang, Yulin and Kang, Bingyi and Han, Yizeng and Wang, Shenzhi and Song, Shiji and Feng, Jiashi and Huang, Gao},
  title     = {{DeeR-VLA}: Dynamic Inference of Multimodal Large Language Models for Efficient Robot Execution},
  booktitle = {Advances in Neural Information Processing Systems},
  year      = {2024},
  url       = {https://openreview.net/forum?id=PBmIq4Z9tq}
}

@inproceedings{zitkovich23a,
  author    = {Zitkovich, Brianna and Yu, Tianhe and Xu, Sichun and Xu, Peng and Xiao, Ted and Xia, Fei and Wu, Jialin and Wohlhart, Paul and Welker, Stefan and Wahid, Ayzaan and others},
  title     = {{RT-2}: Vision-Language-Action Models Transfer Web Knowledge to Robotic Control},
  booktitle = {Proceedings of the Conference on Robot Learning},
  series    = {Proceedings of Machine Learning Research},
  volume    = {229},
  publisher = {PMLR},
  year      = {2023},
  pages     = {2165--2183}
}

@inproceedings{zhang2025citynavagent,
  author    = {Zhang, Weichen and Gao, Chen and Yu, Shiquan and Peng, Ruiying and Zhao, Baining and Zhang, Qian and Cui, Jinqiang and Chen, Xinlei and Li, Yong},
  title     = {{CityNavAgent}: Aerial Vision-and-Language Navigation With Hierarchical Semantic Planning and Global Memory},
  booktitle = {Proceedings of the 63rd Annual Meeting of the Association for Computational Linguistics},
  year      = {2025},
  pages     = {31292--31309},
  doi       = {10.18653/v1/2025.acl-long.1511}
}

@article{mandi2024roco,
  author  = {Mandi, Zhao and Jain, Shreeya and Song, Shuran},
  title   = {{RoCo}: Dialectic Multi-Robot Collaboration With Large Language Models},
  journal = {CoRR},
  volume  = {abs/2307.04738},
  year    = {2023},
  url     = {https://arxiv.org/abs/2307.04738},
  note    = {arXiv:2307.04738}
}

@inproceedings{chen2024scalable,
  author    = {Chen, Yifan and others},
  title     = {Scalable Multi-Robot Collaboration With Large Language Models: Centralized or Decentralized Systems?},
  booktitle = {Proceedings of the IEEE International Conference on Robotics and Automation},
  year      = {2024},
  doi       = {10.1109/ICRA57147.2024.10610855}
}

@inproceedings{ahn2022do,
  author    = {Ahn, Michael and Brohan, Anthony and Brown, Noah and Chebotar, Yevgen and Cortes, Omar and David, Byron and Finn, Chelsea and Fu, Chuyuan and Gopalakrishnan, Keerthana and Hausman, Karol and others},
  title     = {Do As I Can, Not As I Say: Grounding Language in Robotic Affordances},
  booktitle = {Proceedings of the 6th Conference on Robot Learning},
  series    = {Proceedings of Machine Learning Research},
  volume    = {205},
  publisher = {PMLR},
  year      = {2022}
}

@inproceedings{liang2022codeaspolicies,
  author    = {Liang, Jacky and Huang, Wenlong and Xia, Fei and Xu, Peng and Hausman, Karol and Ichter, Brian and Florence, Pete and Zeng, Andy},
  title     = {Code as Policies: Language Model Programs for Embodied Control},
  booktitle = {Proceedings of the IEEE International Conference on Robotics and Automation},
  year      = {2023},
  pages     = {9493--9500},
  doi       = {10.1109/ICRA48891.2023.10160591}
}

@inproceedings{dettmers2024spqr,
  author    = {Dettmers, Tim and Svirschevski, Ruslan A. and Egiazarian, Vage and Kuznedelev, Denis and Frantar, Elias and Ashkboos, Saleh and Borzunov, Alexander and Hoefler, Torsten and Alistarh, Dan-Adrian},
  title     = {{SpQR}: A Sparse-Quantized Representation for Near-Lossless {LLM} Weight Compression},
  booktitle = {Proceedings of the International Conference on Learning Representations},
  year      = {2024},
  url       = {https://openreview.net/forum?id=Q1u25ahSuy}
}

@inproceedings{liu2024spinquant,
  author    = {Liu, Zechun and Zhao, Changsheng and Fedorov, Igor and Soran, Bilge and Choudhary, Dhruv and Krishnamoorthi, Raghuraman and Chandra, Vikas and Tian, Yuandong and Blankevoort, Tijmen},
  title     = {{SpinQuant}: {LLM} Quantization With Learned Rotations},
  booktitle = {Proceedings of the International Conference on Learning Representations},
  year      = {2025},
  url       = {https://openreview.net/forum?id=ogO6DGE6FZ}
}

@inproceedings{liu2024kivi,
  author    = {Liu, Zirui and Yuan, Jiayi and Jin, Hongye and Zhong, Shaochen and Xu, Zhaozhuo and Braverman, Vladimir and Chen, Beidi and Hu, Xia},
  title     = {{KIVI}: A Tuning-Free Asymmetric {2bit} Quantization for {KV} Cache},
  booktitle = {Proceedings of the 41st International Conference on Machine Learning},
  year      = {2024}
}

@inproceedings{shah2024flashattention3,
  author    = {Shah, Jay and Dao, Tri and others},
  title     = {{FlashAttention-3}: Fast and Accurate Attention With Asynchrony and Low Precision},
  booktitle = {Advances in Neural Information Processing Systems},
  year      = {2024},
  url       = {https://arxiv.org/abs/2407.08608}
}

@inproceedings{cai2024medusa,
  author    = {Cai, Tianle and Li, Yuhong and Geng, Zhengyang and Peng, Hongwu and Lee, Jason D. and Chen, Deming and Dao, Tri},
  title     = {{Medusa}: Simple {LLM} Inference Acceleration Framework With Multiple Decoding Heads},
  booktitle = {Proceedings of the 41st International Conference on Machine Learning},
  year      = {2024}
}

@inproceedings{qian2025macnet,
  author    = {Qian, Chen and others},
  title     = {Scaling Large-Language-Model-Based Multi-Agent Collaboration},
  booktitle = {Proceedings of the International Conference on Learning Representations},
  year      = {2025},
  url       = {https://openreview.net/forum?id=K3n5jPkrU6}
}

@article{monwilliams2025embodied,
  author  = {Mon-Williams, Ruaridh and Li, Gen and Long, Ran and Du, Wenqian and Lucas, Christopher G.},
  title   = {Embodied Large Language Models Enable Robots to Complete Complex Tasks in Unpredictable Environments},
  journal = {Nature Machine Intelligence},
  year    = {2025},
  volume  = {7},
  pages   = {592--601},
  doi     = {10.1038/s42256-025-01005-x}
}

@article{10776775,
  author  = {Lou, Jiabin and Shi, Rongye and Lin, Yuxin and Wang, Qunbo and Wu, Wenjun},
  title   = {{TALKER}: A Task-Activated Language Model Based Knowledge-Extension Reasoning System},
  journal = {IEEE Robotics and Automation Letters},
  year    = {2025},
  volume  = {10},
  number  = {2},
  pages   = {1026--1033},
  doi     = {10.1109/LRA.2024.3511434}
}

@article{das2026latencyaware,
  author  = {Das, Murat and Hussain, Zawar and Nawaz, Muhammad},
  title   = {Latency-Aware Benchmarking of Large Language Models for Natural-Language Robot Navigation in {ROS 2}},
  journal = {Sensors},
  year    = {2026},
  volume  = {26},
  number  = {2},
  pages   = {608},
  doi     = {10.3390/s26020608}
}

@inproceedings{zhang2024autoagents,
  author    = {Chen, Guangyao and Dong, Siwei and Shu, Yu and Zhang, Ge and Sesay, Jaward and Karlsson, B{\"o}rje F. and Fu, Jie and Shi, Yemin},
  title     = {{AutoAgents}: A Framework for Automatic Agent Generation},
  booktitle = {Proceedings of the Thirty-Third International Joint Conference on Artificial Intelligence},
  year      = {2024},
  pages     = {21--28},
  doi       = {10.24963/ijcai.2024/3}
}

@inproceedings{zhang2024proagent,
  author    = {Zhang, Ceyao and others},
  title     = {{ProAgent}: Building Proactive Cooperative Agents With Large Language Models},
  booktitle = {Proceedings of the AAAI Conference on Artificial Intelligence},
  year      = {2024},
  volume    = {38},
  number    = {16},
  pages     = {17591--17599},
  doi       = {10.1609/aaai.v38i16.29710}
}

@article{albaroudi2026collabllm,
  author  = {Albaroudi, Elham and others},
  title   = {{COLLAB-LLM}: A Communication-Centric Role-Based Framework for Scalable Multi-Agent {LLM} Collaboration},
  journal = {Asian Journal of Research in Computer Science},
  year    = {2026},
  volume  = {19},
  number  = {1},
  pages   = {152--185},
  doi     = {10.9734/ajrcos/2026/v19i1811}
}

\end{document}